\documentclass[10pt,twocolumn,letterpaper]{article}

\usepackage{cvpr}
\usepackage{times}
\usepackage{epsfig}
\usepackage{graphicx}
\usepackage{amsmath}
\usepackage{amssymb}
\usepackage{caption}
\usepackage{appendix}
\usepackage{cancel}
\usepackage{multirow}
\graphicspath{ {./images/} }

\usepackage[pagebackref=true,breaklinks=true,letterpaper=true,colorlinks,bookmarks=false]{hyperref}

\cvprfinalcopy 

\def\cvprPaperID{7173} 

\ifcvprfinal\fi
\begin{document}

\title{Explorable Super Resolution}

\author{Yuval Bahat and Tomer Michaeli\\
Technion - Israel Institute of Technology, Haifa, Israel\\
{\tt\small \{yuval.bahat@campus,tomer.m@ee\}.technion.ac.il}
}


\twocolumn[{%
	\maketitle
	\vspace{-0.75cm}
	\renewcommand\twocolumn[1][]{#1}%
	\begin{center}
		\centering
		\includegraphics[width=1\textwidth]{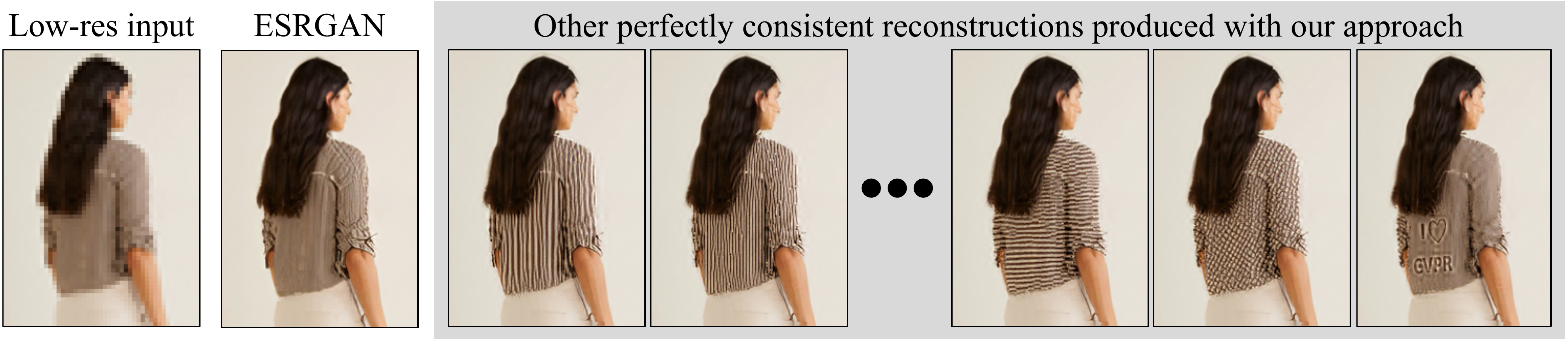}
		\captionof{figure}{
		\textbf{Exploring HR explanations to an LR image.} 
		Existing SR methods (\eg ESRGAN \cite{wang2018esrgan}) output only one explanation to the input image. In contrast, our explorable SR framework allows producing infinite different perceptually satisfying HR images, that all identically match a given LR input, when down-sampled.
		Please zoom-in to view subtle details.
		}
		\label{fig:variable_outputs}
	\end{center}%
}]
\begin{abstract}
Single image super resolution (SR) has seen major performance leaps in recent years. 
However, existing methods do not allow exploring the infinitely many plausible reconstructions that might have given rise to the observed low-resolution (LR) image. These different explanations to the LR image may dramatically vary in their textures and fine details, and may often encode completely different semantic information. 
In this paper, we introduce the task of \emph{explorable super resolution}. We propose a framework comprising a graphical user interface with a neural network backend, allowing editing the SR output so as to explore the abundance of plausible HR explanations to the LR input. At the heart of our method is 
a novel module that can wrap any existing SR network, analytically guaranteeing that its SR outputs would precisely match the LR input, when down-sampled. Besides its importance in our setting, this module is guaranteed to decrease the reconstruction error of any SR network it wraps, and can be used to cope with blur kernels that are different from the one the network was trained for. We illustrate our approach in a variety of use cases, ranging from medical imaging and forensics, to graphics.  

\end{abstract}
\section{Introduction}\label{sec:intro}
\begin{figure*}[t]
\centering
\includegraphics[width=\textwidth]{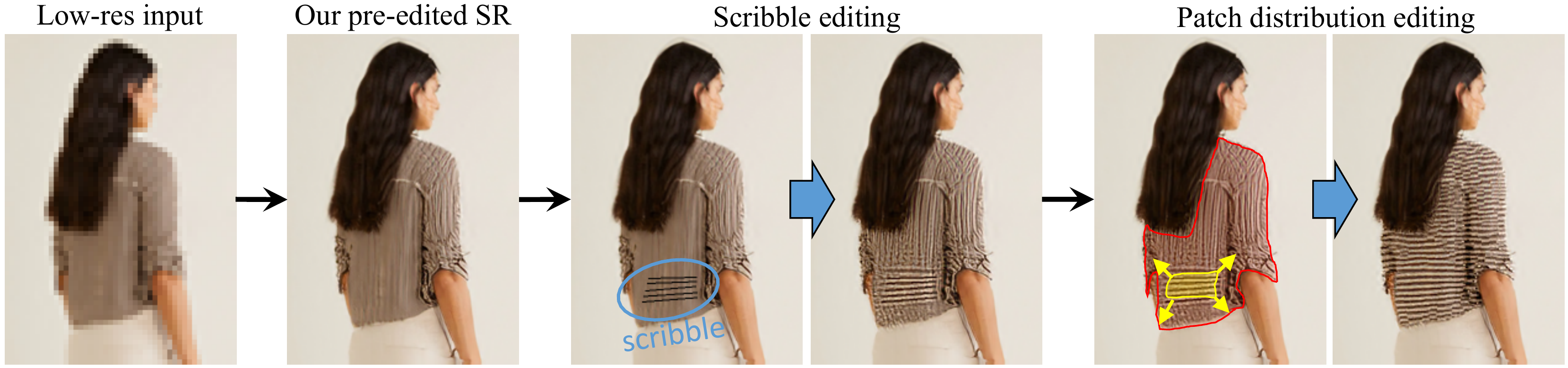}
\caption{
\textbf{An example user editing process.}
Our GUI allows exploring the space of plausible SR reconstructions using a variety of tools. Here, local scribble editing is used to encourage the edited region to resemble the user's graphical input. Then the entire shirt area (red) is edited by encouraging its patches to resemble those in the source (yellow) region.
At any stage of the process, the output is \emph{perfectly consistent} with the input (its down-sampled version identically matches the input). 
}
\label{fig:editing_fig1}
\end{figure*}
Single image \emph{super resolution} (SR) is the task of producing a high resolution (HR) image from a single low resolution (LR) image. Recent decades have seen an increasingly growing research interest in this task, peaking with the recent surge of methods based on deep neural networks. These methods demonstrated significant performance boosts, some in terms of achieving low reconstruction errors \cite{dong2014learning,kim2016accurate,lim2017edsr,shocher2018zssr,lai2018Lap_srn,Zhang_2018_denseNet_sr,kligvasser2018xunit} and some in terms of producing photo-realistic HR images \cite{ledig2017srgan,wang2018esrgan,shaham2019singan}, typically via the use of generative adversarial networks (GANs) \cite{goodfellow2014gans}. 
However, common to all existing methods is that they do not allow exploring the abundance of plausible HR explanations to the input LR image, and typically produce only a \emph{single} SR output. This is dissatisfying as although these HR explanations share the same low frequency content, manifested in their coarser image structures, they may significantly vary in their higher frequency content, such as textures and small details (see \eg, Fig.~\ref{fig:variable_outputs}). Apart from affecting the image appearance, these fine details often encode crucial semantic information, like in the cases of text, faces and even textures (\eg, distinguishing a horse from a zebra). Existing SR methods ignore this abundance of valid solutions, and arbitrarily confine their output to a specific appearance with its particular semantic meaning.

\begin{figure}[!t]
\centering
\includegraphics[width=\columnwidth]{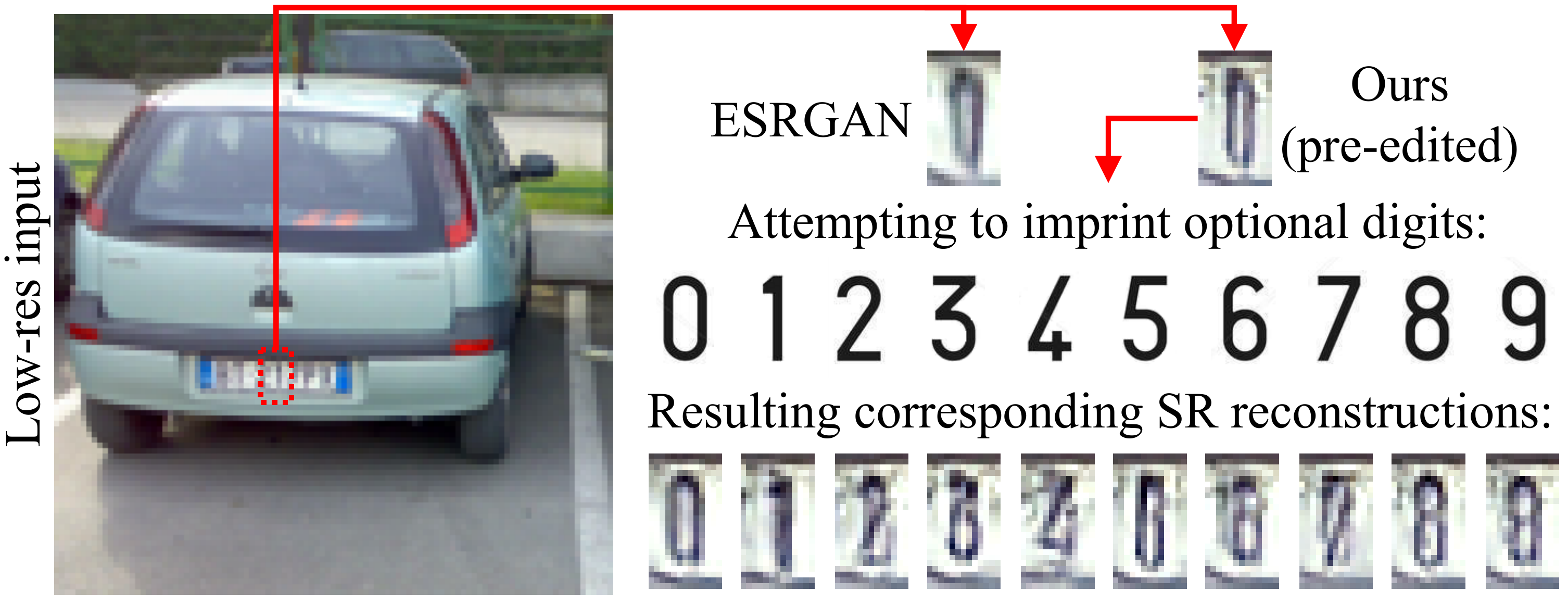}
\vspace{-18pt}\caption{\label{fig:license_plate}\textbf{Visually examining the likelihood of patterns of interest.} Given an LR image of a car license plate, we explore the possible valid SR reconstructions by attempting to manipulate the central digit to appear like any of the digits $0-9$, using our imprinting tool (see Sec.~\ref{sec:editing_tools}). Though the ground truth HR digit was $1$, judging by the ESRGAN \cite{wang2018esrgan} result (or by our pre-edited reconstruction) would probably lead to misidentifying it as $0$. In contrast, our results when imprinting digits $0$,$1$ and $8$ contain only minor artifacts, thus giving them similar likelihood. 
}
\end{figure}

\begin{figure}[!t]
\centering
\includegraphics[width=\columnwidth]{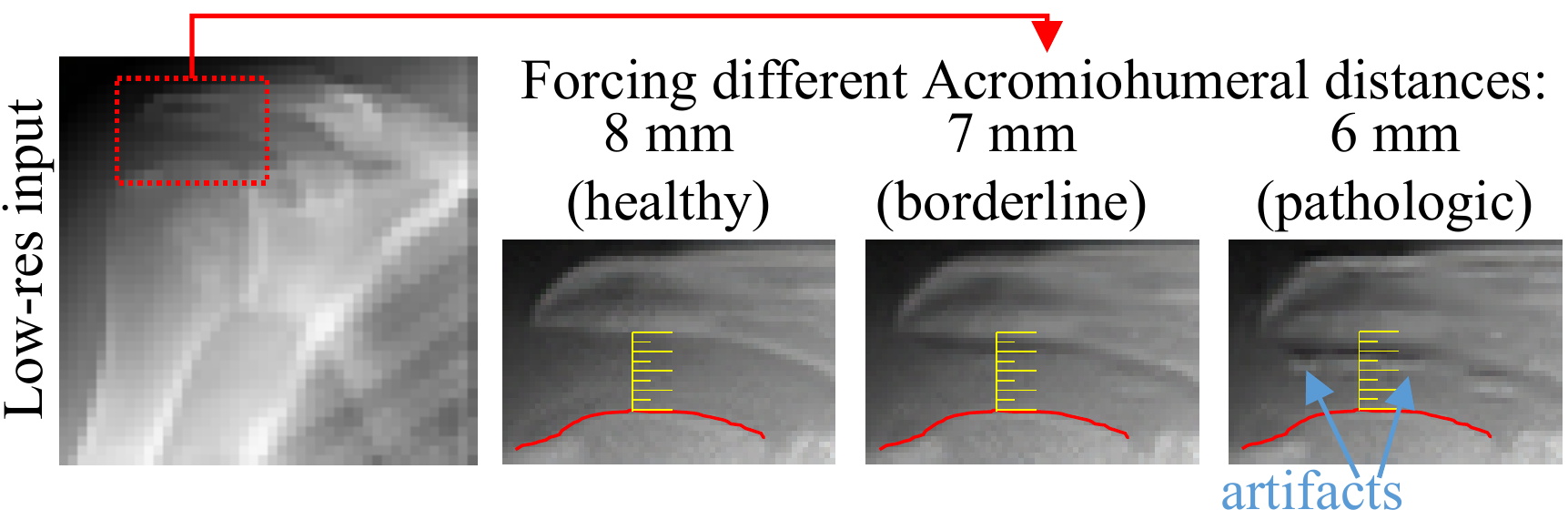}
\vspace{-18pt}\caption{\label{fig:Xray}\textbf{Visually examining the likelihood of a medical pathology.} Given an LR shoulder X-ray image, we evaluate the likelihood of a Supraspinatus tendon tear, typically characterized by a less than 7mm Acromiohumeral distance (measured between the Humerus bone, marked red, and the Acromion bone above it). To this end, we attempt to imprint down-shifted versions (see Sec.~\ref{sec:editing_tools}) of the Acromion bone. Using the image quality as a proxy for its plausibility, we can infer a low chance of pathology, due to the artifacts emerging when forcing the pathological form (right image).
}
\end{figure}

In this paper, we initiate the study of \emph{explorable super resolution}, and propose a framework for achieving it through user editing. Our method consists of a neural network utilized by a graphical user interface (GUI), which allows the user to interactively explore the space of perceptually pleasing HR images that could have given rise to a given LR image. An example editing process is shown in Fig.~\ref{fig:editing_fig1}. 
Our approach is applicable in numerous scenarios. For example, it enables manipulating the image so as to fit any prior knowledge the user may have on the captured scene, like changing the type of flora to match the capturing time and location, adjusting shades according to the capturing time of day, or manipulating an animal's appearance according to whether the image was taken in the zebras or horses habitat. 
It can also help determine whether a certain pattern or object could have been present in the scene. This feature is invaluable in many settings, including in the forensic and medical contexts, exemplified in Figs.~\ref{fig:license_plate} and~\ref{fig:Xray}, respectively. Finally, it may be used to correct unpleasing SR outputs, which are common even with high capacity neural network models.

\begin{figure}[t]
\centering
\includegraphics[width=\columnwidth]{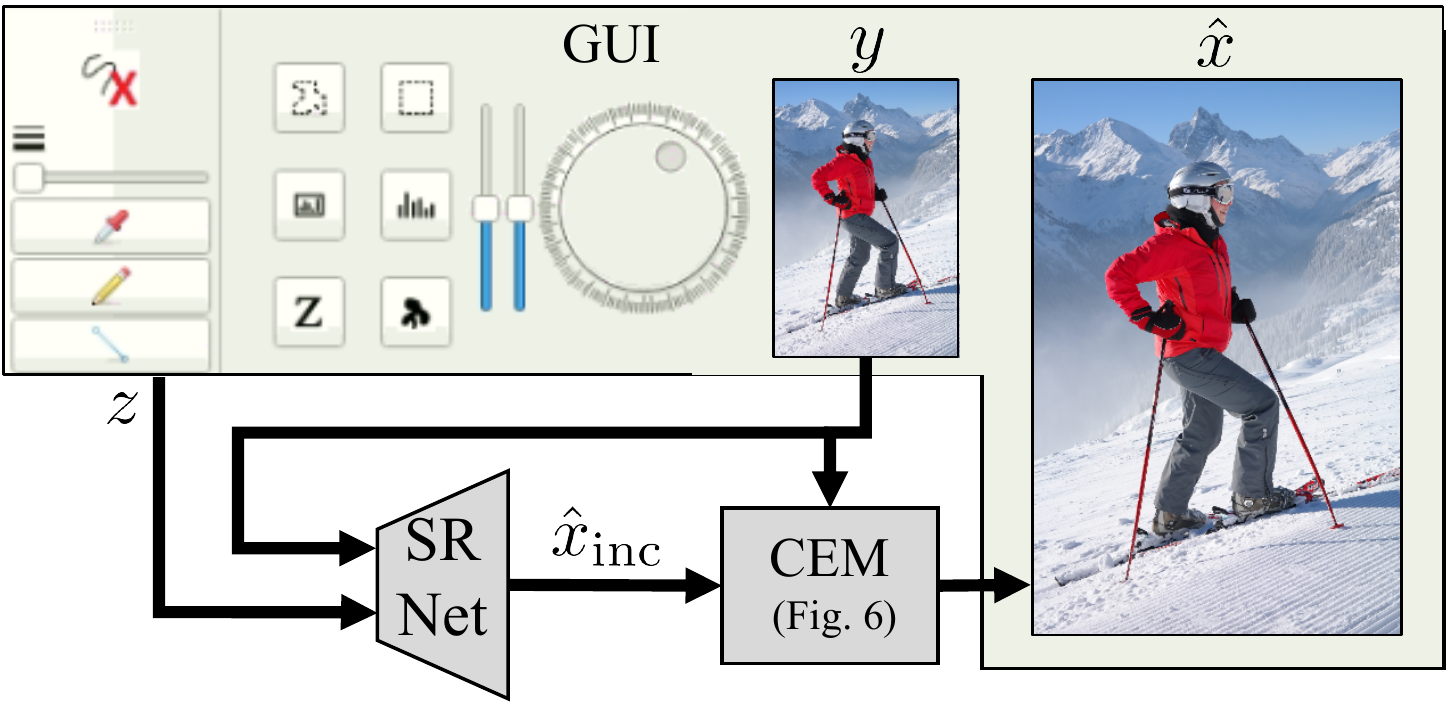}
\vspace{-18pt}\caption{\label{fig:framework_scheme}\textbf{Our explorable SR framework.} Our GUI allows interactively exploring the manifold of possible HR explanations for the LR image $y$, by manipulating the control signal $z$ of the SR network. Our CEM is utilized to turn the network's output $\hat{x}_{\text{inc}}$ into a consistent reconstruction $\hat{x}$, presented to the user. See Secs.~\ref{sec:consistency_module},\ref{sec:editable_SR_net} and~\ref{sec:editing_tools} for details.}
\end{figure}

Our framework (depicted in Fig.~\ref{fig:framework_scheme}) consists of three key ingredients, which fundamentally differ from the common practice in SR. (i)~We present a novel \emph{consistency enforcing module} (CEM) that can wrap any SR network, analytically guaranteeing that its outputs identically match the input, when down-sampled. Besides its crucial role in our setting, we illustrate the advantages of incorporating this module into any SR method. (ii)~ We use a neural network with a control input signal, which allows generating diverse HR explanations to the LR image. To achieve this, we rely solely on an adversarial loss to promote perceptual plausibility, without using \emph{any reconstruction loss} (\eg $L_1$ or VGG) for promoting proximity between the network's outputs and the ground-truth HR images.
(iii)~We facilitate the exploration process by creating a GUI comprising a \emph{large set of tools}. These work by manipulating the network's control signal so as to achieve various desired effects\footnote{Our code and GUI are available online.}. We elaborate on those three ingredients in Secs.~\ref{sec:consistency_module},\ref{sec:editable_SR_net} and~\ref{sec:editing_tools}.

\subsection{Related Work}
\paragraph{GAN based image editing}
Many works employed GANs for image editing tasks. For example, Zhu \etal~\cite{zhu2016gan_editing} 
performed editing by searching for an image that satisfies a user input scribble, while constraining the output image to lie on the natural image manifold, learned by a GAN. Perarnau \etal~\cite{perarnau2016invertible_editing} suggested to perform editing in a learned latent space, by combining an encoder with a conditional GAN. Xian \etal~\cite{Xian_2018_texture_editing} facilitated texture editing, by allowing users to place a desired texture patch. Rott Shaham \etal~\cite{shaham2019singan} trained their GAN solely on the image to be edited, thus encouraging their edited output to retain the original image characteristics. While our method also allows traversing the natural image manifold, it is different from previous approaches in that it enforces the hard consistency constraint (restricting all outputs to identically match the LR input when down-sampled).

\paragraph{GAN based super-resolution} 
A large body of work has shown the advantage of using \emph{conditional GANs} (cGANs) for generating photo-realistic SR reconstructions \cite{ledig2017srgan,wang2018esrgan,sajjadi2017enhancenet,wang2018realistic_texture,wang2018progressive}. 
Unlike classical GANs, cGANs feed the generator with additional data (\eg an LR image) together with the stochastic noise input. The generator then learns to synthesize new data (\eg the corresponding SR image) conditioned on this input. In practice, though, cGAN based SR methods typically feed their generator only with the LR image, without a stochastic noise input. Consequently, they do not produce diverse SR outputs. 
While we also use a cGAN, we do add a control input signal to our generator's LR input, which allows editing its output to yield diverse results.
Several cGAN methods for image translation did target outputs' diversity by keeping the additional stochastic input \cite{zhu2017toward_multimodal,chen2016infogan,karras2019stylegan}, while utilizing various mechanisms for binding the output to this additional input. In our method, we encourage diversity by simply removing the reconstruction losses that are used by all existing SR methods. This is made possible by our consistency enforcing module.
\section{The Consistency Enforcing Module}
\label{sec:consistency_module}
We would like the outputs of our explorable SR method to be both perceptually plausible and consistent with the LR input. 
To encourage perceptual plausibility, we adopt the common practice of utilizing an adversarial loss, which penalizes for deviations from the statistics of natural images. To guarantee consistency, we introduce the \emph{consistency enforcing module} (CEM), an architecture that can wrap any given SR network, making it inherently satisfy the consistency constraint. This is in contrast to existing SR networks, which do not perfectly satisfy this constraint, as they encourage consistency only \emph{indirectly} through a reconstruction loss on the SR image. The CEM does not contain any learnable parameters and has many notable advantages over existing SR network architectures, on which we elaborate later in this section.
We next derive our module.

Assume we are given a low resolution image $y$ that is related to an unknown high-resolution image $x$ through
\begin{equation}
    \label{eq:downscale_conv}
    y=(h*x)\downarrow_\alpha.
\end{equation}
Here, $h$ is a blur kernel associated with the point-spread function of the camera, $*$ denotes convolution, and $\downarrow_\alpha$ signifies sub-sampling by a factor~$\alpha$. With slight abuse of notation, \eqref{eq:downscale_conv} can be written in matrix form as
\begin{equation}
    \label{eq:downscale_matrix}
    y = H x,
\end{equation}
where $x$ and $y$ now denote the vectorized versions of the HR and LR images, respectively, and the matrix $H$ corresponds to convolution with $h$ and sub-sampling by~$\alpha$.
This system of equations is obviously under-determined, rendering it impossible to uniquely recover $x$ from $y$ without additional knowledge. We refer to any HR image $\hat{x}$ satisfying this constraint, as \emph{consistent} with the LR image $y$. We want to construct a module that can project any inconsistent reconstruction $\hat{x}_{\text{inc}}$ (\eg the output of a pre-trained SR network) onto the affine subspace defined by \eqref{eq:downscale_matrix}. Its consistent output $\hat{x}$ is thus the minimizer of
\begin{equation}\label{eq:projH}
\min_{\hat{x}} \|\hat{x}-\hat{x}_{\text{inc}}\|^2 \quad \text{s.t.}\quad H\hat{x}=y.
\end{equation}
Intuitively speaking, such a module would guarantee that the low frequency content of $\hat{x}$ matches that of the ground-truth image $x$ (manifested in $y$), so that the SR network should only take care of plausibly reconstructing the high frequency content (\eg sharp edges and fine textures).

Problems like \eqref{eq:projH} frequently arise in sampling theory (see \eg~\cite{michaeli2010optimization}), and can be conveniently solved using a geometric viewpoint.
Specifically, let us utilize the fact that \mbox{$P_{\mathcal{N}(H)^\perp}=H^T(HH^T)^{-1}H$} is known to be the orthogonal projection matrix onto $\mathcal{N}(H)^\perp$, the subspace that is perpendicular to the nullspace of $H$. Now, multiplying both sides of the constraint in~\eqref{eq:projH} by $H^T(HH^T)^{-1}$, yields
\begin{equation}\label{eq:projNperp}
P_{\mathcal{N}(H)^\perp}\hat{x}=H^T(HH^T)^{-1}y.
\end{equation}
This shows that we should strictly set the component of $\hat{x}$ in $\mathcal{N}(H)^\perp$ to equal the right hand side of \eqref{eq:projNperp}.

We are therefore restricted to minimize the objective by manipulating only the complementary component,
$P_{\mathcal{N}(H)}\hat{x}$, that lies in the nullspace of $H$. Decomposing the objective into the two subspaces, 
\begin{equation}
\|P_{\mathcal{N}(H)}(\hat{x}-\hat{x}_{\text{inc}})\|^2+\|P_{\mathcal{N}(H)^\perp}(\hat{x}-\hat{x}_{\text{inc}})\|^2,
\end{equation}
we see that $P_{\mathcal{N}(H)}\hat{x}$ only appears in the first term, which it minimizes when set to
\begin{equation}\label{eq:projN}
    P_{\mathcal{N}(H)}\hat{x}=P_{\mathcal{N}(H)}\hat{x}_{\text{inc}}.
\end{equation}
Combining the two components from \eqref{eq:projNperp} and \eqref{eq:projN}, and using the fact that $P_{\mathcal{N}(H)}=I-H^T(HH^T)^{-1}H$, we get that
\begin{align}
\label{eq:full_s_matrix}
\hat{x}&=P_{\mathcal{N}(H)}\hat{x} + P_{\mathcal{N}(H)^\perp}\hat{x} \nonumber\\
&= (I-H^T(HH^T)^{-1}H)\hat{x}_{\text{inc}} + H^T(HH^T)^{-1}y.
\end{align}

\begin{figure}[t]
\centering
\includegraphics[width=\columnwidth]{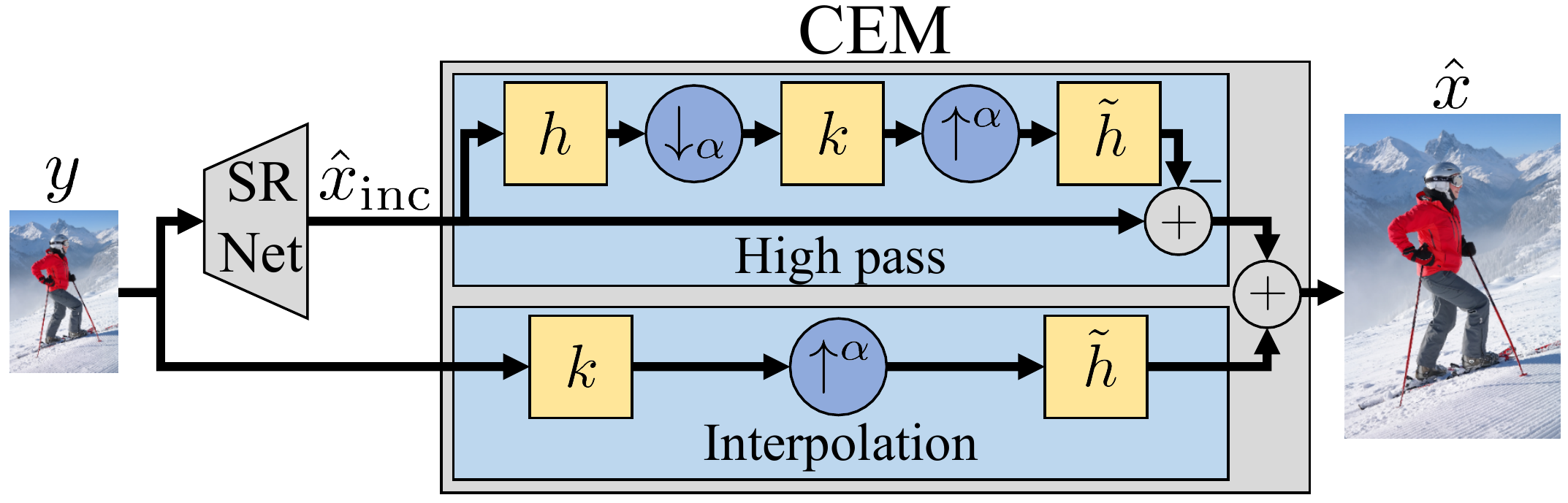}
\vspace{-18pt}\caption{\label{fig:architechture}\textbf{CEM architecture.} The CEM, given by Eq.~\eqref{eq:full_s_conv}, can wrap any given SR network. 
It projects its output $\hat{x}_{\text{inc}}$ onto the space of images that identically match input $y$ when downsampled, thus producing a super-resolved image $\hat{x}$ guaranteed to be consistent with $y$.
See Sec.~\ref{sec:consistency_module} for details.
}
\end{figure}

To transform \eqref{eq:full_s_matrix} into a practical module that can wrap any SR architecture with output $\hat{x}_{\text{inc}}$, we need to replace the impractical multiplication operations involving the very large $H$ matrix, with their equivalent operations: convolutions, downsampling and upsampling. To this end, we observe that since $H$ corresponds to convolution with $h$ followed by sub-sampling, $H^T$ corresponds to up-sampling followed by convolution with a mirrored version of $h$, which we denote by $\tilde{h}$. 
The multiplication by $(HH^T)^{-1}$ can then be replaced by convolving with a filter $k$, constructed by computing the inverse of the filter $(h*\tilde{h})\downarrow_\alpha$ in the Fourier domain. 
We thus have that
\begin{equation}
    \label{eq:full_s_conv}
    \hat{x}=\hat{x}_{\text{inc}}-\tilde{h}*\big[k*(h*\hat{x}_{\text{inc}})\downarrow_\alpha\big]\uparrow^\alpha + \tilde{h}*(k*y)\uparrow^\alpha.
\end{equation}
Thus, given the blur kernel\footnote{We default to the bicubic blur kernel when $h$ is not given.} $h$, we can calculate the filters appearing in \eqref{eq:full_s_conv} and hardcode their non-learned weights into our CEM, which can wrap any SR network, as shown in Fig.~\ref{fig:architechture} (see Supplementary 
for padding details). Before proceeding to incorporate it in our scheme, we note
the CEM is beneficial for \emph{any} SR method, in the following two aspects.

\begin{table}
    \centering
    \begin{tabular}{|c|c|c|c|}
        \hline
        &$2\times$&$3\times$&$4\times$\\
        \hline
        EDSR \cite{lim2017edsr}&35.97&32.27&30.30\\
        \hline
        EDSR+CEM&\textbf{36.11}&\textbf{32.36}&\textbf{30.37} \\
        \hline
    \end{tabular}
    \caption{\textbf{Wrapping a pre-trained SR network with CEM.} Mean PSNR values over the BSD100 dataset \cite{martin2001bsd100}, super-resolved by factors $2,3$ and $4$. Merely wrapping the network with our CEM can only improve the reconstruction error, as manifested by the slight PSNR increase in the 2\textsuperscript{nd} row.}
    \label{tab:wrapping_EDSR}
\end{table}{}

\vspace{-10pt}
\paragraph{Reduced reconstruction error}
Merely wrapping a pre-trained SR network with output $\hat{x}_{\text{inc}}$ by our CEM, can only decrease its reconstruction error w.r.t.~the ground-truth $x$, as
\begin{align}
\|\hat{x}_{\text{inc}}-x\|^2 \geq \|P_{\mathcal{N}(H)}(\hat{x}_{\text{inc}}-x)\|^2 
&\overset{(\text{a})}{=}\|P_{\mathcal{N}(H)}(\hat{x}-x)\|^2 \nonumber \\
&\overset{(\text{b})}{=} \|\hat{x}-x\|^2.
\end{align}
Here, (a) is due to \eqref{eq:projN}, and (b) follows from \eqref{eq:projNperp}, which implies that $\|P_{\mathcal{N}(H)^\perp}(\hat{x}-x)\|^2=0$ (as $P_{\mathcal{N}(H)^\perp}\hat{x}=H^T(HH^T)^{-1}Hx=P_{\mathcal{N}(H)^\perp}x$). This is demonstrated in Tab.~(\ref{tab:wrapping_EDSR}), which reports PSNR values attained by the EDSR network \cite{lim2017edsr} with and without our CEM, for several different SR scale factors.

\begin{figure}[!t]
\centering
\includegraphics[width=\columnwidth]{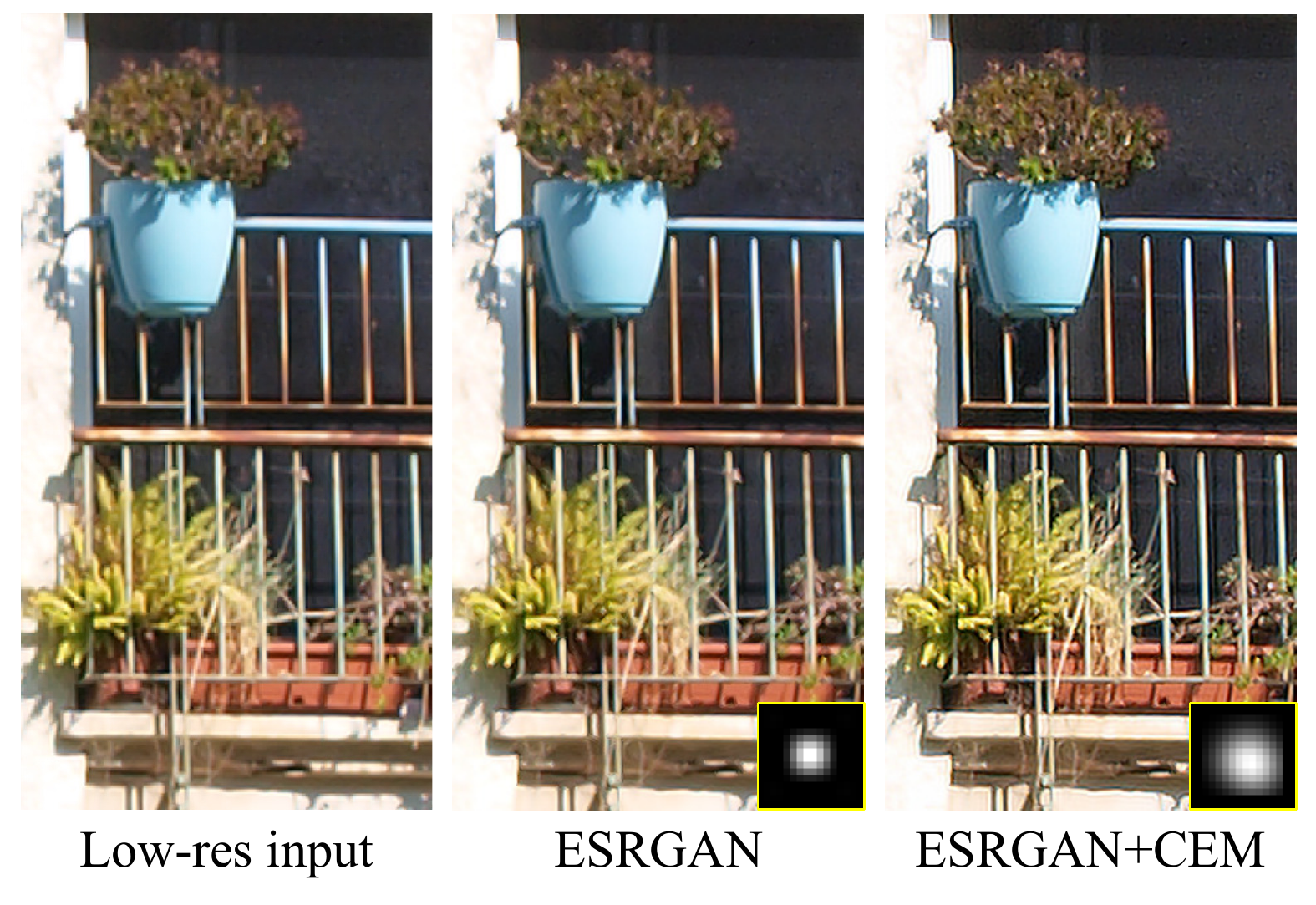}
\vspace{-18pt}\caption{\label{fig:estimated_kernel}\textbf{Accounting for a different blur kernel at test time.} The CEM can be used to adopt any SR network to a blur kernel other than the one it was trained on.
We demonstrate this on a real LR image. Here, ESRGAN \cite{wang2018esrgan}, which was trained with the bicubic kernel, produces blurry results. However, wrapping it with our CEM, with the kernel estimated by \cite{bell2019kernel_gan}, results in a much sharper reconstruction. Corresponding kernels are visualized on each image.
}
\end{figure}

\vspace{-10pt}
\paragraph{Adopting to a non-default down-sampling kernel}
Deep learning based SR methods are usually trained on LR images obtained by down-sampling HR images using a specific down-sampling kernel (usually the bicubic kernel). This constitutes a major drawback, as their performance significantly degrades when processing LR images corresponding to different kernels, as is the case with most realistic images \cite{Michaeli2013BlindSR,shocher2018zssr}. This problem can be alleviated using our CEM, that takes the down-sampling kernel as a parameter upon its assembly at test time. Thus, it can be used for adopting a given network that was pre-trained using some default kernel, to the kernel corresponding to the LR image. Figure~\ref{fig:estimated_kernel} demonstrates the advantage of this approach on a real LR image, using the kernel estimation method of~\cite{bell2019kernel_gan}.

\section{Editable SR Network}\label{sec:editable_SR_net}
To achieve our principal goal of creating an explorable super resolution network, we develop a new framework, that comprises a GUI with a neural network backend (see Fig.~\ref{fig:framework_scheme}).
Our SR network differs from existing SR methods in two core aspects. First, it is capable of producing a \emph{wide variety} of plausible and inherently consistent HR reconstructions $\hat{x}$, for any input LR image $y$. This is achieved thanks to the CEM, which omits the need to use any reconstruction loss (\eg $L_1$ or VGG) for driving the outputs close on average to the corresponding ground truth training images. Such losses bias the outputs towards the average of all possible explanations to the LR image, and are thus not optimal for the purpose of exploration. Second, our network incorporates a control input signal $z$, that allows traversing this manifold of plausible images so as to achieve various manipulation effects on the output $\hat{x}$.
%
Using our GUI, the user can either tune $z$ manually to achieve simple effects (\eg control the orientation and magnitude of gradients as in Fig.~\ref{fig:direct_Z_editing}) or use any of our automatic manipulation tools that take as input \eg user scribbles or some pattern to imprint (see Figs.~\ref{fig:editing_fig1}-\ref{fig:Xray},\ref{fig:tools},\ref{fig:animal_variations}).

We use the same network architecture as ESRGAN \cite{wang2018esrgan},
but train it wrapped by the CEM, and by minimizing a loss function comprising four terms,
\begin{equation}
    \label{eq:total_loss}
    {\cal L}_\text{Adv}+\lambda_\text{Range}{\cal L}_\text{Range}+\lambda_{\text{Struct}}{\cal L}_{\text{Struct}}+\lambda_\text{Map}{\cal L}_\text{Map}.
\end{equation}
Here, ${\cal L}_\text{Adv}$ is an adversarial loss, which encourages the network outputs to follow the statistics of natural images. We specifically use a Wasserstein GAN loss with gradient penalty \cite{gulrajani2017wgan_gp}, and avoid using the relativistic discriminator \cite{jolicoeur2018relativistic} employed in ESRGAN, as it induces a sort of full-reference supervision.
The second loss term, ${\cal L}_\text{Range}$, penalizes for pixel values that exceed the valid range $[0,1]$, and thus helps prevent model divergence. We use ${\cal L}_\text{Range}=\tfrac{1}{N}\|\hat{x}-\text{clip}_{[0,1]}\{\hat{x}\}\|_1$, where $N$ is the number of pixels. The last two loss terms, ${\cal L}_\text{Struct}$ and ${\cal L}_\text{Map}$, are associated with the control signal $z$.
We next elaborate on the control mechanism and these two penalties.

\subsection{Incorporating a Control Signal}\label{subsec:effective_control}
As mentioned above, to enable editing the output image $\hat{x}$, we introduce a control signal $z$, which we feed to the network in addition to the input image $y$. 
We define the control signal as $z\in \mathbb{R}^{w\times h\times c}$, where $w\times h$ are the dimensions of the output image $\hat{x}$ and $c=3$, to allow intricate editing abilities (see below). To prevent the network from ignoring this additional signal, as reported for similar cases in \cite{pix2pix2016,mathieu2015video_prediction}, we follow the practice in \cite{navarrete2018multi} and concatenate $z$ to the input of each layer of the network, where layers with smaller spatial dimensions are concatenated with a spatially downscaled version of $z$.
At test time, we use this signal to traverse the space of plausible HR reconstructions. Therefore, at train time, we would like to encourage the network to associate different $z$ inputs to different HR explanations. To achieve this, we inject random $z$ signals during training. 

\begin{table}[t]
    \centering
    \begin{tabular}{c|c|c|c|}
        \cline{2-4}
        &Diversity&Percept.&Reconst.\\
        &&quality&error\\
        &$(\sigma)$&(NIQE)&(RMSE)\\
        \hline
        \multicolumn{1}{|c|}{ESRGAN}
        &$0$&$3.5\pm0.9$&$17.3\pm7.2$\\
        \hline
        \multicolumn{1}{|c|}{ESRGAN with $z$}
        &$3.6\pm1.7$&$3.7\pm0.8$&$17.5\pm6.9$\\
        \hline
        \multicolumn{1}{|c|}{Ours}
        &$7.2\pm3.4$&$3.7\pm0.9$&$18.2\pm7.4$\\  
        \hline
    \end{tabular}
    \caption{\textbf{Quality and diversity of SR results.} We report diversity (standard deviation, higher is better), perceptual quality (NIQE \cite{mittal2012niqe}, lower is better) and RMSE (lower is better), for $4\times$ SR on the BSD100 test set \cite{martin2001bsd100}. 
    Values are measured over 50 different SR outputs per input image, produced by injecting 50 random, spatially uniform $z$ inputs. 
    Note that our model, trained without any full-reference loss terms, shows a significant advantage in terms of diversity, while exhibiting similar perceptual quality. 
    See Supplementary for more details about this experiment.} 
    \label{tab:losses_ablation}
\end{table}{}

\begin{figure*}[t]
\centering
\includegraphics[width=\textwidth]{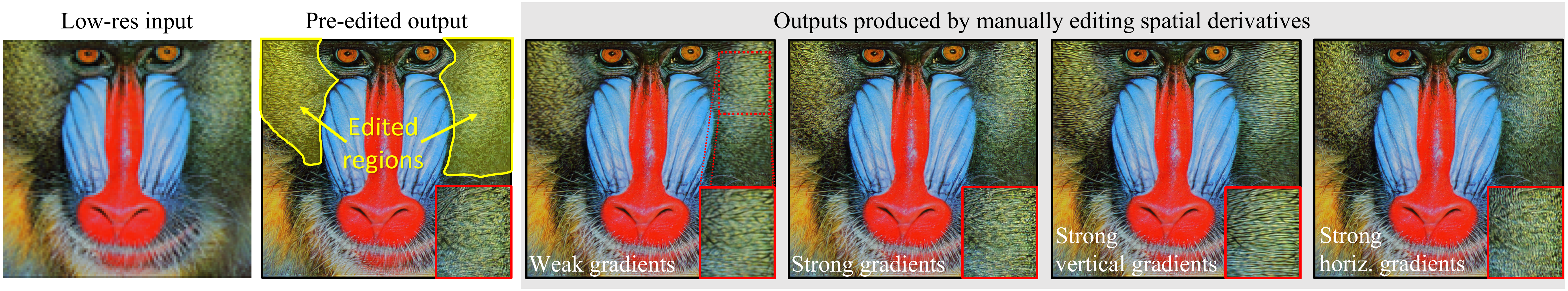}
\vspace{-18pt}\caption{\label{fig:direct_Z_editing}\textbf{Manipulating image gradients.} A user can explore different perceptually plausible textures (\eg the Mandrill's cheek fur), by manually adjusting the directions and magnitudes of image gradients 
using the control signal $z$. 
}
\end{figure*}

Incorporating this input signal into the original ESRGAN method already affects outputs diversity. This can be seen in Tab.~\ref{tab:losses_ablation}, which compares the vanilla ESRGAN method (1\textsuperscript{st} row) with its variant, augmented with $z$ as described above (2\textsuperscript{nd} row), both trained for additional $6000$ generator steps using the original ESRGAN loss. 
However, we can obtain an even larger diversity. Specifically, recall that as opposed to the original ESRGAN, in our loss we use no reconstruction (full-reference) penalty that resists diversity.
The effect can be seen in the 3\textsuperscript{rd} row of Tab.~\ref{tab:losses_ablation}, which corresponds to our model trained for the same number of steps\footnote{Weights corresponding to $z$ in the 2\textsuperscript{nd} and 3\textsuperscript{rd} rows' models are initialized to $0$, while all other weights are initialized with the pre-trained ESRGAN model from the 1\textsuperscript{st} row.} using only the ${\cal L}_\text{Adv}$ and ${\cal L}_\text{Range}$ loss terms.
Note that all three models are on par in terms of perceptual quality.

Now that the output of our network strongly depends on $z$, we move on to make this dependence easy to manipulate by a user. To allow manual control, we want simple modifications to $z$ to lead to variations that are intuitive to grasp. To allow optimization-based manipulations, we want to ensure that any plausible output image could be generated by some $z$. These two requirements are encouraged by the last two loss terms in \eqref{eq:total_loss}, as we describe next.

\vspace{-10pt}
\paragraph{Controlling image gradients with the ${\cal L}_{\text{Struct}}$ loss}
To allow intuitive manual control, we encourage spatially uniform perturbations to $z$ to affect the spatial derivatives of the output image $\hat{x}$.
Figure~\ref{fig:direct_Z_editing} demonstrates how manipulating the gradients' magnitudes and directions, can be used to edit textures, like fur. We choose to associate the 3 channels of $z$ with the 3 degrees of freedom of the structure tensor of $\hat{x}$, which is the $2\times2$ matrix defined by ${\cal S}_{\hat{x}}=\iint (\nabla \hat{x}(\xi,\eta)) (\nabla \hat{x}(\xi,\eta))^Td\xi d\eta$. 
We encode this link into our network through the loss term ${\cal L}_{\text{Struct}}$, which penalizes for the difference between desired structure tensors ${\cal S}_\text{d}$, determined by a randomly sampled, spatially uniform $z$, and the tensors corresponding to the actual network's outputs, ${\cal S}_{\hat{x}}$ (see Supplementary for more details).

\begin{figure}[!t]
\centering
\vspace{-5pt}
\includegraphics[width=0.6\columnwidth]{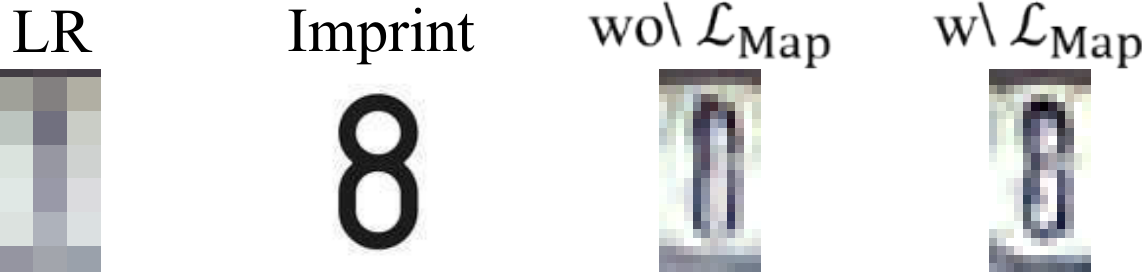}
\caption{\label{fig:Lmap_loss}\textbf{Effect of ${\cal L}_\text{Map}$.} Incorporating the ${\cal L}_\text{Map}$ loss term into our model's training introduces spatially non-uniform $z$ inputs, and improves the model's ability to cover the natural image manifold. We demonstrate this by comparing a model trained with vs.~without this term, in the task of imprinting the digit `8' on the license plate of Fig.~\ref{fig:license_plate}.
}
\end{figure}

\vspace{-10pt}
\paragraph{Facilitating optimization based editing via the ${\cal L}_{\text{Map}}$ loss}
Denoting our network output by $\hat{x}=\psi(y,z)$, we would like to guarantee that $\psi$ can generate every plausible image $\hat{x}$ with some choice of $z$. To this end, we introduce the loss term ${\cal L}_\text{Map}=\min_z\|\psi(y,z)-x\|_1$, which penalizes for differences between the real natural image $x$, and its best possible approximation using some signal $z$. 
Within each training step, we first solve the internal minimization of ${\cal L}_\text{Map}$ over $z$ for $10$ iterations, and then freeze this $z$ for the minimization of all loss terms in~\eqref{eq:total_loss}. 
Note that in contrast to ${\cal L}_{\text{Struct}}$, which only involves spatially uniform $z$ inputs, the minimization over $z$ in
${\cal L}_\text{Map}$
exposes our network to the entire $z$ space during training, encouraging its mapping onto the entire natural image manifold. Figure~\ref{fig:Lmap_loss} illustrates how incorporating the ${\cal L}_\text{Map}$ loss term improves our model's ability to generate plausible patterns.

\subsection{Training details}
We train our network using the $800$ DIV2K training images \cite{agustsson2017div2k}. 
We use the standard bicubic downsampling kernel,
both for producing LR training inputs and 
as $h$ in our CEM (Eq.~\eqref{eq:full_s_conv}), and feed our network with $64\times64$ input patches, randomly cropped from these LR images. We initialize our network weights with the pre-trained ESRGAN \cite{wang2018esrgan} model, except for weights corresponding to input signal $z$, which we initialize to zero. 
We minimize the loss in \eqref{eq:total_loss} 
with $\lambda_\text{Range}=5000,\lambda_{\text{Struct}}=1$ and $\lambda_\text{Map}=100$.
We set the Wasserstein GAN gradient penalty weight to $\lambda_\text{gp}=10$.
We establish critic's credibility before performing each generator
step, by verifying that the critic correctly distinguishes fake images from real ones for 10 consecutive batches.
We use a batch size of $48$ and train for $\sim80K$ steps.
\section{Editing Tools}\label{sec:editing_tools}
\begin{figure*}[!t]
\centering
\includegraphics[width=\textwidth]{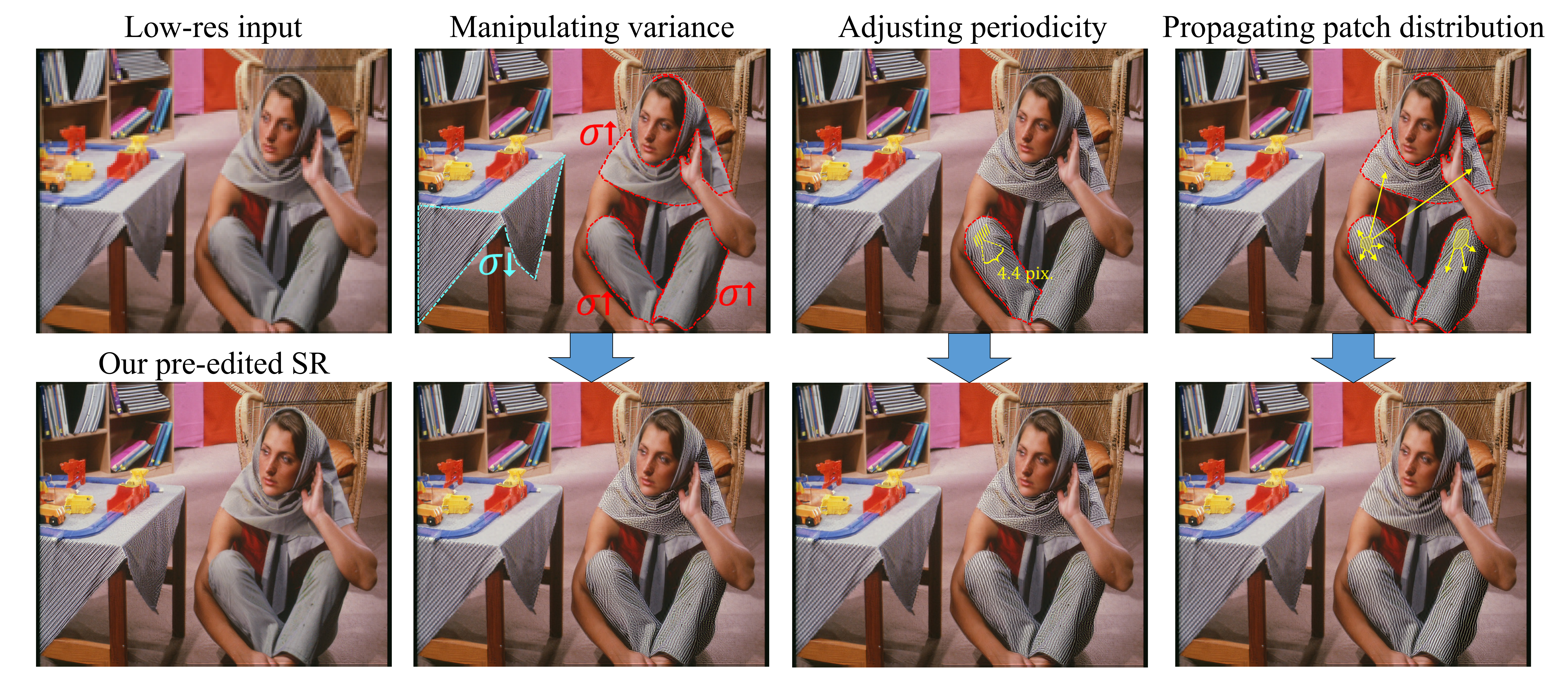}
\caption{\label{fig:tools}\textbf{Editing using tools that optimize over $\bf{z}$.} Here we use our variance manipulation, periodicity, and patch dictionary tools. Each optimizes over $z$ space to achieve a different objective. See Sec.~\ref{sec:editing_tools} for more details.}
\end{figure*}
\begin{figure*}[!t]
\centering
\includegraphics[width=\textwidth]{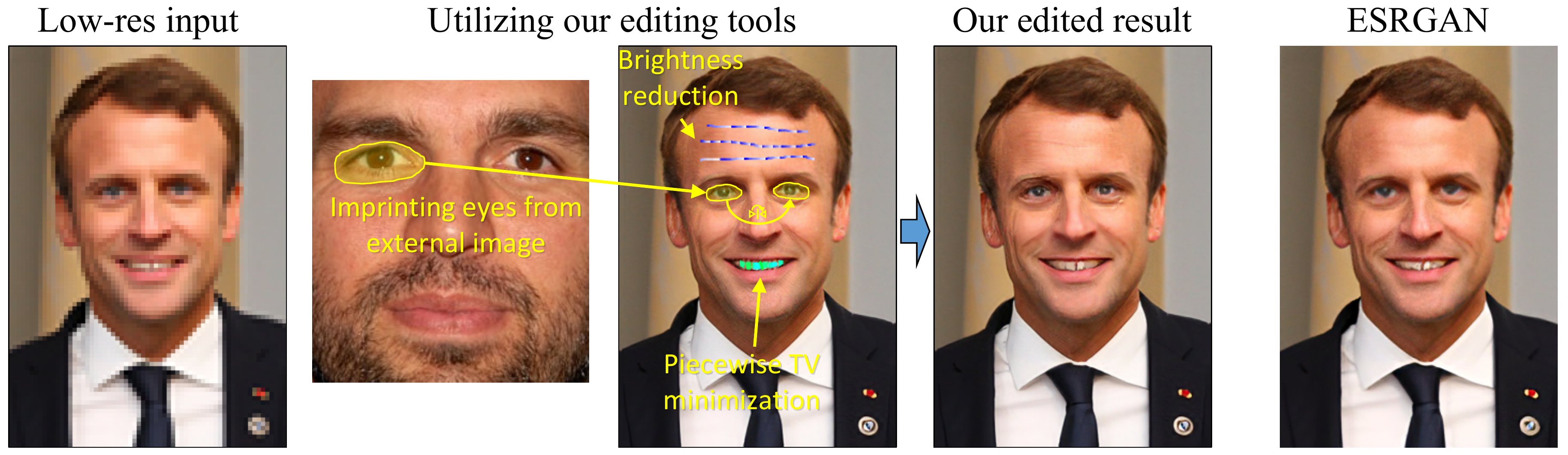}
\caption{\label{fig:face_editing}\textbf{Utilizing scribble editing tools.} Our tools allow regaining small details and producing crisper and more realistic images (compared to, \eg ESRGAN \cite{wang2018esrgan}). Here we use the piecewise smoothness tool (TV minimization) to enhance edges between teeth, and restore forehead wrinkles using our brightness reduction pen. We imprint the eye region from an online available image of (brown-eyed) Mark Ruffalo to restore the crisp appearance of the blue eyes of Emmanuel Macron.}
\end{figure*}

We incorporate our trained SR network described in Sec.~\ref{sec:editable_SR_net} as the backend engine of an editing GUI, that allows manipulating the output SR image $\hat{x}$ by properly shaping the control signal $z$. 
Our GUI comprises several tools that can be applied on selected regions of the image, which we achieve by manipulating only the corresponding regions of~$z$ (recall that the dimensions of~$z$ and $\hat{x}$ are the same).

The most basic tool constitutes three knobs (top middle in Fig.~\ref{fig:framework_scheme}) that manually control the triple channeled $z$ signal in a spatially uniform manner, so as to affect the image structure tensor ${\cal S}_{\hat{x}}$ (Sec.~\ref{subsec:effective_control}). We make this tool user intuitive by binding the knobs with the eigenvalue decomposition of ${\cal S}_{\hat{x}}$, thus affecting image gradients. The user can control the orientation $\theta$ and magnitude $\lambda_1$ of prominent gradients, and the magnitude $\lambda_2$ in the perpendicular direction (see Fig.~\ref{fig:direct_Z_editing}). 

To allow a more diverse set of intricate editing operations, we also propose tools that optimize specific editing objectives (\eg increasing local image variance). These tools invoke a gradient descent optimization process over~$z$, whose goal is to minimize the chosen objective. This is analogous to traversing the manifold of perceptually plausible valid SR images (captured by our network), while remaining consistent with the LR image (thanks to the CEM). 

Our GUI recomputes $\hat{x}$ after each edit in nearly real-time, ranging from a couple of milliseconds for the basic tools to 2-3 seconds for editing a $100\times 100$ region using the~$z$ optimization tools (with an NVIDIA GeForce 2080 GPU). 
We next briefly survey and demonstrate these tools, 
and leave the detailed description of each objective to the supplementary.

\vspace{-10pt}\paragraph{Variance manipulation}
This set of tools searches for a $z$ that decreases or increases the variance of pixel values. This is demonstrated in Fig.~\ref{fig:tools} for yielding a smoother table map and more textured trousers. A variant of this tool allows \emph{locally} magnifying or attenuating exiting patterns.

\vspace{-10pt}\paragraph{User scribble}
Our GUI incorporates a large set of tools for imposing a graphical input by the user on the SR image. A user first scribbles on the current SR image and chooses the desired color and line width. Our GUI then optimizes over $z$, searching for the image $\hat{x}$ that satisfies the imposed input while lying on the learned manifold of valid perceptually plausible SR images. Figure~\ref{fig:editing_fig1} (1\textsuperscript{st} editing operation) demonstrates how we use line scribbles for creating the desired shirt pattern. Variants of this tool (demonstrated in Fig.~\ref{fig:face_editing}) enable increasing or decreasing brightness, as well as enforcing smoothness across a specific scribble mark, by minimizing local TV.

\vspace{-10pt}
\paragraph{Imprinting}
This tool enables a user to impose content taken from an external image, or from a different region within the same image. After selecting the desired region, our GUI utilizes the CEM to combine the low frequency content corresponding to $y$ with the high frequency content of the region to be imprinted. This results in a consistent image region, but which may not necessarily be perceptually plausible. We then invoke an optimization over $z$, which attempts to drive $\hat{x}$ close to this consistent region, while remaining on the natural image manifold. Example usages are shown in Figs.~\ref{fig:variable_outputs} (right-most image),~\ref{fig:license_plate},~\ref{fig:face_editing} and~\ref{fig:Xray}, where the tool is used for subtle shifting of image regions.

\vspace{-10pt}\paragraph{Desired dictionary of patches}
This tool manipulates the edited region to comprise patches stemming from a desired patch collection. A user first marks a source region containing desired patches. This region may be taken either from the edited image or from an external image. Our GUI then optimizes $z$ to encourage the target edited region to comprise the same desired patches, where we use patches of size $6\times6$. Figures~\ref{fig:editing_fig1} (2\textsuperscript{nd} editing operation) and \ref{fig:tools} (right) show how we use this tool to propagate a desired local cloth pattern (yellow region) to the entire garment. Variants of this tool allow matching patches while ignoring their mean value or variance, to allow disregarding color and pattern magnitude differences between source and target patches.

\vspace{-10pt}\paragraph{Encouraging periodicity}
This tool enhances the periodic nature of the image along one or two directions marked by the user. The desired period length can be manually adjusted, as exemplified in Fig.~\ref{fig:tools} for the purpose of producing different stripe widths. Alternatively, a user can choose to enhance the most prominent existing period length, automatically estimated by our GUI.

\vspace{-10pt}\paragraph{Random diverse alternatives}
This tool offers an easy way to explore the image manifold. When invoked, it optimizes $z$ to simultaneously produce $N$ different SR image alternatives, by maximizing the $L_1$ distance between them, in pixel space. The user can then use each of the alternatives (or sub-regions thereof) as a baseline for further editing. A variant of this tool constraints the different alternatives to remain close to the current SR image.

\begin{figure}[!t]
\centering
\includegraphics[width=\columnwidth]{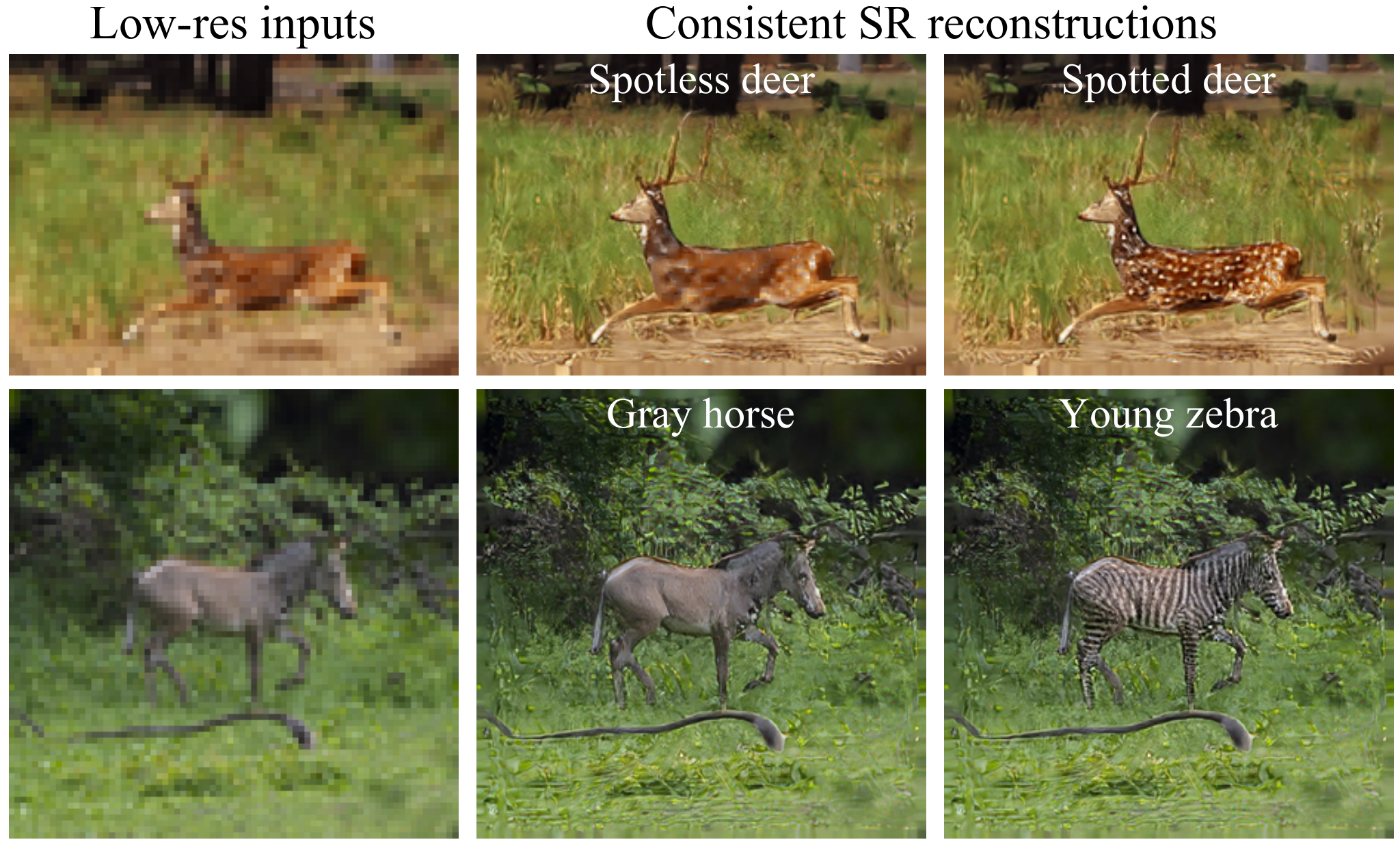}
\vspace{-17pt}\caption{\label{fig:animal_variations}\textbf{More exploration examples.} Using our framework to explore possible SR images corresponding to different semantic contents, all consistent with the LR input.
\vspace{-10pt}}
\end{figure}
\begin{figure}[!t]
\centering
\includegraphics[width=\columnwidth]{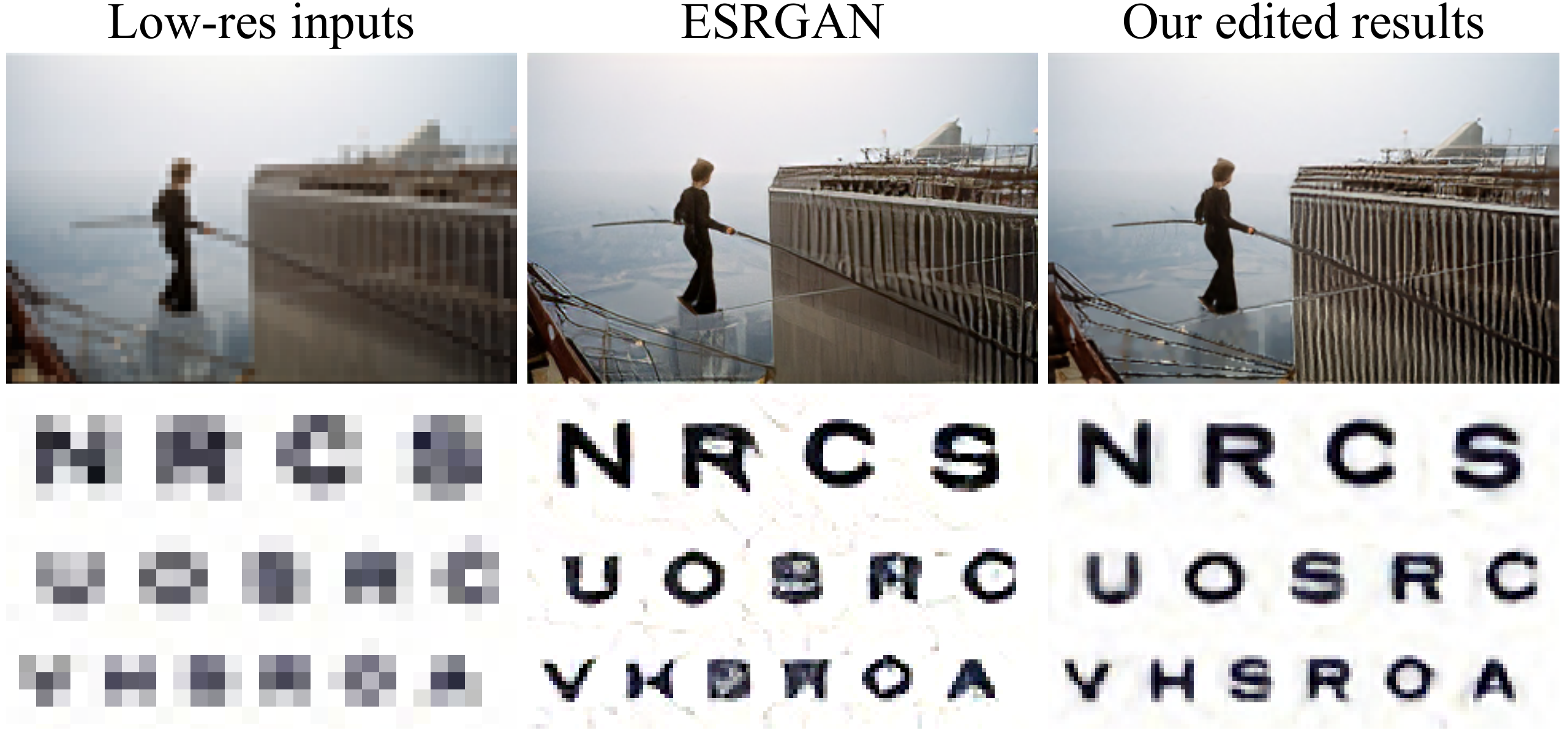}
\vspace{-18pt}\caption{\label{fig:improved_SR}\textbf{Correcting SR outputs.} Our method can be used to enhance SR results, relying on the user's prior knowledge (\eg the appearance of buildings and 
Alphabet characters).
Please zoom-in to view subtle differences, and refer to the Supplementary for the editing processes.
\vspace{-0pt}}
\end{figure}

The wide applicability of our method and editing tools is further demonstrated in Figs.~\ref{fig:animal_variations} and~\ref{fig:improved_SR} in two use cases.
Figure~\ref{fig:animal_variations} demonstrates our method's exploration capabilities, illustrating how
utterly different semantic content can be generated for the same LR input.
Figure~\ref{fig:improved_SR} shows how our method's editing capabilities can enhance SR outputs, by relying on the user's external knowledge.
Please see more editing process examples and results in Supplementary.

\vspace{-0pt}
\section{Conclusion}\label{sec:conclusion}
A low resolution image may correspond to many different HR images. However, existing SR methods generate only a single, arbitrarily chosen, reconstruction. In this work, we introduced a framework for explorable super resolution, which allows traversing the set of perceptually plausible HR images that are consistent with a given LR image. We illustrated how our approach is beneficial in many domains, including in medicine, surveillance, and graphics.



\vspace{-10pt}
\paragraph{Acknowledgment}
This research was supported by the Israel Science Foundation (grant 852/17) and by the Technion Ollendorff Minerva Center.

   {
   \newpage
   \twocolumn[
   \null
   \vskip .375in
   \begin{center}
      {\Large \bf Explorable Super Resolution - Supplementary Material \par}
      \vspace*{24pt}
      {
      \large
      \lineskip .5em
      \begin{tabular}[t]{c}
         Yuval Bahat and Tomer Michaeli\\
Technion - Israel Institute of Technology, Haifa, Israel\\
{\tt\small \{yuval.bahat@campus,tomer.m@ee\}.technion.ac.il}\\
         \vspace*{1pt}\\
      \end{tabular}
      \par
      }
      \vskip .5em
      \vspace*{12pt}
   \end{center}
   ]
}
\appendix
\section{Boundary Effects and Padding}\label{app:boundary_effects}
Our consistency enforcing module comprises several filter convolution operations (see Fig.~\ref{fig:architechture}), each resulting in reduced output size. We follow the common practice of zero-padding filters' outputs to match their input sizes, which causes the consistency guarantee to brake, as we get closer to image boundaries, potentially resulting in undesired boundary effects.

To prevent these boundary effects from affecting the training of our editable SR network, we avoid utilizing outer image regions when calculating our loss function \eqref{eq:total_loss}, by first removing the $10\times\alpha$ outer pixels of our network's output, where $\alpha$ is the SR scale factor. In inference mode, we reduce boundary effects by replicating the LR image boundary pixels $10$ times, before feeding it to our network, then removing the $10\times\alpha$ outer pixels from the network's output. This works better than the common zero-padding.
\section{Diversity Comparison Experiment (Tab.~\ref{tab:losses_ablation})}\label{app:ablation_study}
We performed an experiment to compare the effect of incorporating control input signal $z$ and omitting reconstruction loss terms on the diversity and quality of SR outputs. The results in Tab.~\ref{tab:losses_ablation} indicate that while incorporating $z$ into the original ESRGAN model already facilitates producing diverse outputs, training using our scheme, that uses no reconstruction loss (thanks to the CEM) yields a significant boost in outputs diversity. This is while all compared models demonstrate similar perceptual quality. We next present details about the experiment setup and results processing.

\subsection{Experimental Setup}
We study the case of $4\times$ SR and initialize weights of all three models in the table using the official pre-trained ESRGAN model \cite{wang2018esrgan}, where weights corresponding to the $z$ signal (in the 2\textsuperscript{nd} and 3\textsuperscript{rd} rows' models) are initialized to zero, and train each model for additional 6000 generator steps on the DIV2K dataset \cite{agustsson2017div2k}. Models in first two rows are trained using the original ESRGAN loss function, that includes full-reference reconstruction loss terms, namely $L_1$ and VGG. Even the adversarial loss is computed relative to a reference ground truth image, as they employ a relativistic discriminator \cite{jolicoeur2018relativistic}. Our model in the 3\textsuperscript{rd} row is trained using only the ${\cal L}_{\text{Adv}}$ and ${\cal L}_{\text{Range}}$ loss terms, without using any full reference terms.

We evaluate all three models using the 100 BSD100 dataset \cite{martin2001bsd100} images. For the 2\textsuperscript{nd} and 3\textsuperscript{rd} models we produce $50$ different versions of $\hat{x}$, by running the model $50$ times for each input image $y$, each time with a different, randomly sampled, spatially uniform $z$.

\subsection{Metrics Calculation}
The first row's model has 0 diversity, as it can only output one SR output per LR input. To evaluate diversity for the other two models, we calculate the per-pixel standard deviation across the different $z$ inputs, and average over all image pixels and color channels, yielding a per input image value $\sigma_i,~i=1\dots100$. The diversity and error margin values presented for each model in Tab.~\ref{tab:losses_ablation} correspond to the average and standard deviation of $\sigma_i$ over all $100$ dataset images, respectively.
However, recall from Sec.~\ref{sec:consistency_module} that since our method's outputs are consistent with their corresponding input image $y$, their diversity can only be manifested in the component of $\hat{x}$ lying in the nullspace of $H$, $P_{\mathcal{N}(H)}\hat{x}$. We therefore project $\hat{x}$ onto this nullspace \emph{before} performing the diversity calculations described above for both latter models, to yield a more relevant diversity score.
To put the presented diversity values in context, note that the average spatial standard deviation of $P_{\mathcal{N}(H)}x$ (the component of the real HR image $x$ lying in $H$'s nullspace) is $15.2$.

We use the NIQE score \cite{mittal2012niqe} to evaluate perceptual quality. We compute the score for each image, and for the latter two models average over all image versions. The perceptual quality score and error margins for each model are then calculated by calculating the mean and standard deviation over the entire BSD100 dataset, respectively. To put the presented NIQE scores into context, note that the average score for the real HR images is $3.1$.

Finally, we use the Root Mean Square Error (RMSE) measure to quantify reconstruction error. For each output image $\hat{x}$, we measure the square differences between $\hat{x}$ and its corresponding ground truth image $x$, and then average over all image pixels and over the three color channels. Taking the square root yields a per image RMSE. We then follow the same procedure as with the former two metrics to yield the average score and error margins for each model.

\section{Utilizing Loss Term ${\cal L}_{\text{Struct}}$}\label{app:structure_tensor_loss}
As we explain in Sec.~\ref{subsec:effective_control}, we encourage control signal $z$ to affect the magnitude and direction of spatial derivatives of output image $\hat{x}$, by associating $z$ with the structure tensor of $\hat{x}$, denoted ${\cal S}_{\hat{x}}$. The structure tensor is a $2\times2$ matrix computed by
\mbox{${\cal S}_{\hat{x}}=\iint (\nabla \hat{x}(\xi,\eta)) (\nabla \hat{x}(\xi,\eta))^Td\xi d\eta$}, where $\nabla \hat{x}(\xi,\eta)$ is the local gradient of image $\hat{x}$ in pixel $(\xi,\eta)$.
We encode this link into our network through the loss term ${\cal L}_{\text{Struct}}$, which penalizes for the difference between the structure tensors corresponding to the network's outputs, ${\cal S}_{\hat{x}}$, and desired structure tensors ${\cal S}_\text{d}$, determined by a randomly sampled, spatially uniform $z$, as we explain next.

For each training image in each optimization batch, we perform the following steps:
\begin{enumerate}
    \item Drawing random desired structure tensors ${\cal S}_\text{d}$:
    \begin{enumerate}
        \item We uniformly sample $\lambda_{1,2}\in[0,1]$ and $\theta\in[0,2\pi]$, corresponding to desired magnitudes and direction of prominent image edges, respectively. These parameters induce a specific \emph{singular value decomposition} (SVD) of an image structure tensor, and therefore uniquely determine it.
        \item\label{item:structure_tensor_SVD} We then compose the induced structure tensor as \mbox{\hspace{-20pt}
        ${\cal S}_\text{d}=\begin{bmatrix}\lambda_1\cos^2{\theta}+\lambda_2\sin^2{\theta} & \lambda_1\lambda_2\sin{\theta}\cos{\theta} \\ \lambda_1\lambda_2\sin{\theta}\cos{\theta} & \lambda_1\sin^2{\theta}+\lambda_2\cos^2{\theta} \end{bmatrix}$}.
    \end{enumerate}
    \item Feeding and running our model:
    \begin{enumerate}
        \item We set the input control signal $z$ to be spatially constant. Its three channels are set to the values on and above the diagonal of ${\cal S}_\text{d}$, normalized to lie in a symmetric range around zero. 
        \item We feed $z$ along with the image $y$ into our model, and compute the resulting image $\hat{x}$.
        \item We calculate the corresponding structure tensor of $\hat{x}$,
        ${\cal S}_{\hat{x}}$.
    \end{enumerate}
    \item The plausible magnitudes range of spatial derivatives varies according to the local nature of the image, \eg images containing prominent textures correspond to larger magnitudes than those corresponding to smooth images. It therefore makes sense to grant the user control over relative magnitudes rather than over absolute ones, by adjusting the values of ${\cal S}_{\hat{x}}$ and ${\cal S}_\text{d}$ as following:
    \begin{enumerate}
        \item We adjust ${\cal S}_{\hat{x}}$ by computing the sum over the magnitude of the product of the horizontal and vertical spatial image derivatives in the corresponding ground truth image $x$, and use this value to normalize ${\cal S}_{\hat{x}}$, yielding ${\cal \tilde{S}}_{\hat{x}}$.
        \item We adjust each value ${{\cal S}_\text{d}}[i,j]$ in ${\cal S}_\text{d}$ to lie in $[P_5({\cal \tilde{S}}_{\hat{x}}[i,j]),P_{95}({\cal \tilde{S}}_{\hat{x}}[i,j])]$, the $5^{\text{th}}$ and $95^{\text{th}}$ percentiles of ${\cal \tilde{S}}_{\hat{x}}[i,j]$, respectively, collected over 10K images:
        \begin{align*}
        {{\cal \tilde{S}}_\text{d}}[i,j]\triangleq & \frac{P_{95}({{{\cal \tilde{S}}_{\hat{x}}}[i,j]})-P_{5}({{{\cal \tilde{S}}_{\hat{x}}}[i,j]})}{2}{{\cal S}_\text{d}}[i,j] \nonumber\\
        &+ \frac{P_{95}({{{\cal \tilde{S}}_{\hat{x}}}[i,j]})+P_{5}({{{\cal \tilde{S}}_{\hat{x}}}[i,j]})}{2}.
        \end{align*}
    \end{enumerate}
    \item Finally, we calculate the per-image loss as the difference between ${{\cal \tilde{S}}_{\hat{x}}}$ and ${{\cal \tilde{S}}_\text{d}}$:
    \begin{align*}
        {\cal L}_{\text{Struct}}=
        |{{{\cal \tilde{S}}_{\hat{x}}}[1,1]}
        -{{\cal \tilde{S}}_\text{d}}[1,1]|
        &+|{{{\cal \tilde{S}}_{\hat{x}}}[1,2]}-{{\cal \tilde{S}}_\text{d}}[1,2]|\nonumber\\
        &+|{{{\cal \tilde{S}}_{\hat{x}}}[2,2]}-{{\cal \tilde{S}}_\text{d}}[2,2]|.
    \end{align*}
\end{enumerate}

\section{Editing Tools}
Our GUI comprises many editing and exploration tools. The most basic tool allows a user to adjust the local spatial image derivatives, as demonstrated in Fig.~\ref{fig:direct_Z_editing}, by tuning three knobs corresponding to the three channels of $z$. To allow more intricate operations, we propose additional tools that optimize specific editing objectives (\eg increasing local image variance). These tools invoke a gradient descent optimization process over $z$, whose goal is to minimize the chosen objective. This is analogous to traversing the manifold of perceptually plausible valid SR images (captured by our network), while always remaining consistent with the LR image (thanks to the CEM). After briefly surveying these tools in Sec.~\ref{sec:editing_tools}, we next elaborate in detail on each one of them. Tools operating on image patches (rather than directly on the image itself) use partially overlapping $6\times6$ patches. The degree of overlap varies, and indicated separately for each tool.

\subsection{Variance Manipulation}
This set of tools operates by manipulating the local variance of all partially overlapping image patches in the selected region. We optimize over $z$ to increase or decrease the per-patch variance by a pre-determined given user value. 
\paragraph{Signal magnitude manipulation}
A variant of this tool preserves patches' structures and only manipulates the magnitude of their signal, after removing their mean values. This variant operates on less patches, by sub-sampling overlapping patches using a 4 rows stride.
\subsection{User Scribble}
Our GUI constitutes a large set of Microsoft-Paint-like tools, allowing a user to impose a graphical input on the output image. These include pen and straight line tools (with adjustable line width), as well as polygon, square and circle drawing tools. Scribbling color can be chosen manually or sampled from any given image (including the edited one). After scribbling, the user initiates the optimization process, traversing the $z$ space in an attempt to find the image $\hat{x}$ that is closest to the desired user input, while lying on the learned manifold of valid, perceptually plausible images.
\paragraph{Brightness manipulation}
A user can use a variant of all scribble tools to increase or decrease the current local image brightness by a given, adjustable, factor. To this end, the user chooses the brightness increase/decrease ``color'' rather than a standard color, when drawing the graphical input. Once initiated, the optimization process of this variant tool attempts to satisfy the desired local relative brightness changes, rather than absolute graphical inputs.
\paragraph{Local TV minimization}
Similar to the brightness manipulation variant, this variant operates by separately minimizing the total variations (TV) in each separate scribble input. Choosing the TV minimization ``color'', the user can use any scribble tool to mark desired regions. The optimization process would then operate on all pixels in each marked region, attempting to minimize the magnitude of the differences between each pixel and its $8$ neighboring pixels.
\subsection{Imprinting}
This tool enables a user to enforce graphical content on the SR output. The user first selects the desired content, either from within the edited image or from an external one. The user then marks a bounding rectangle on the edited image, to which the desired content is imprinted. The imprinting itself uses the CEM module to keep the original low frequency from $y$, while manipulating only the high frequency content using the desired input. This already guarantees the imprinted region is consistent with the LR image $y$.  The user can then modify the exact location and size of imprinted content using $4$ arrow buttons. Finally, optimizing over the $z$ space searches for the closest image $\hat{x}$ that is not only consistent, but also lies on the learned manifold of natural images.
\paragraph{Subtle region shifting}
Instead of marking the desired imprinting location and size, a user can choose to imprint content taken from the edited image onto itself. The user can then utilize the $4$ arrow buttons to move the selected content from its original location and modify its size, thus inducing subtle shifting of selected image regions.
\subsection{Desired Dictionary of Patches}
This tool manipulates (target) patches in a desired region to resemble the patches comprising a desired (source) region, either taken from an external image or from a different location in the edited image. To allow encouraging desired textures across regions with different colors, we first remove mean patch values from each patch, in both source and target patches. To reduce computational load, we discard some of the overlapping patches, by using $2$ and $4$ rows strides in the source and target regions, respectively. Once source and target regions are selected, optimizing over $z$ traverses the learned manifold of valid plausible images to minimize the distance between each target patch and its closest source patch.
\paragraph{Ignoring patches' variance}
A variant of this tool allows encouraging desired textures without changing current local variance. To this end, we normalize patches' variance, in addition to removing their mean. Then while optimizing over $z$, we add an additional objective to preserve the original variance of each target patch, while encouraging its (normalized) signal to resemble that of its closest (normalized) source patch.
\subsection{Encouraging Periodicity}
This tool encourages the periodic nature of an image region, across one or two directions determined by a user. The desired period length (in pixels) for each direction can be manually set by the user, or it can be automatically set to the most prominent period length, by calculating local image self-correlation. Periodicity is then encouraged by penalizing for the difference between the image region and its version translated by a single period length, for each desired direction.
\subsection{Random Diverse Alternatives}
As we describe in Sec.~\ref{sec:editing_tools}, this tool allows exploring the image manifold, producing $N$ different SR outputs by maximizing the $L_1$ distance between them in pixel space. These images (or sub-regions thereof) can then serve as a baseline for further editing and exploration.
\paragraph{Constraining distance to current image}
This variant adds the requirement that all $N$ images should be close to the current $\hat{x}$, in terms of $L_1$ distance in pixel space.
\section{Additional Results and Editing Processes}
\begin{figure*}[!t]
\centering
\includegraphics[width=\textwidth]{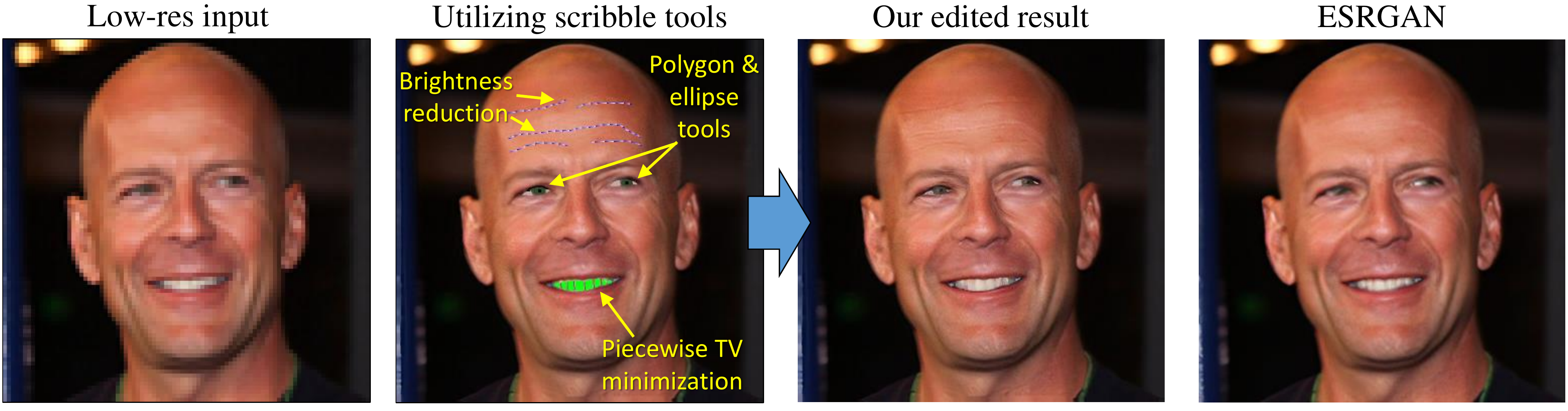}
\caption{\label{fig:editing_Bruce}\textbf{Face editing using only the scribble tool.} Editing a face image of Bruce Willis, utilizing only the scribble tools and its variants, brightness reduction and local TV minimization tools.}
\end{figure*}

\begin{figure*}[!t]
\centering
\includegraphics[width=\textwidth]{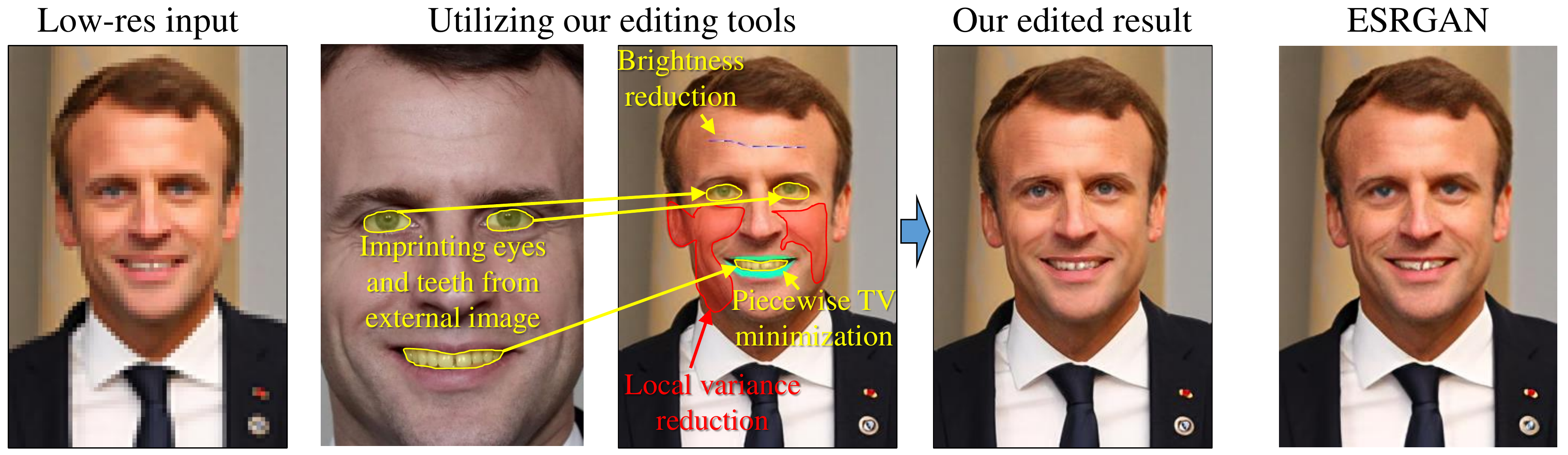}
\caption{\label{fig:face_editing_Macron_using_Macron}\textbf{An alternative to the editing process in Fig.~\ref{fig:face_editing}.} We use our imprinting tool to imprint the high resolution versions of Macron's eyes and teeth. Unlike the case in Fig.~\ref{fig:face_editing}, having access to another image of the subject at hand can yield enhanced results. Other tools utilized in this example are piecewise TV minimization, local variance reduction and local brightness reduction.}
\end{figure*}

\begin{figure*}[!t]
\centering
\includegraphics[width=\textwidth]{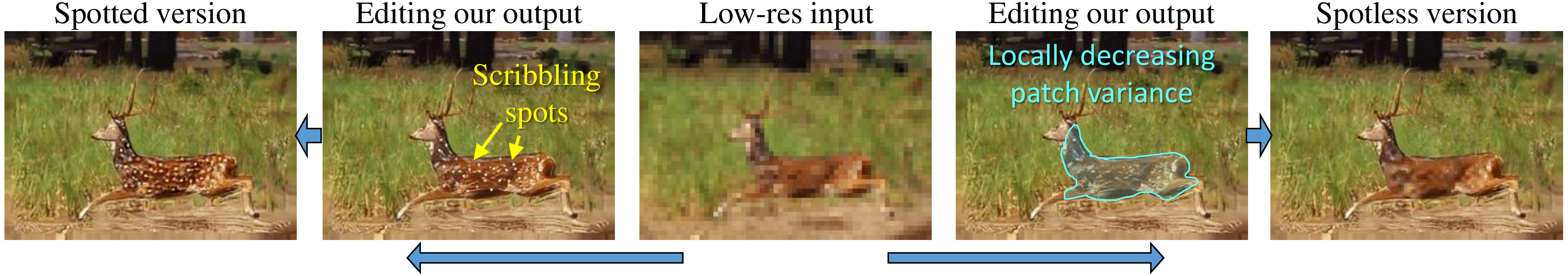}
\caption{\label{fig:editing_spotted_deer}\textbf{The making of Fig.~\ref{fig:animal_variations} (top).} Exploring possible SR solutions to the LR input, by performing two different editing processes, yielding two different deer species.}
\end{figure*}

\begin{figure*}[!t]
\centering
\includegraphics[width=\textwidth]{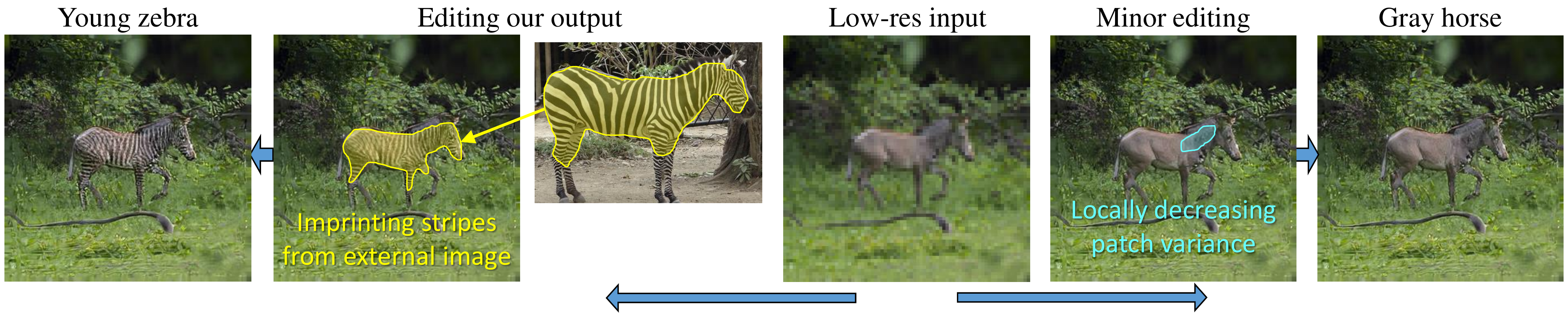}
\caption{\label{fig:editing_zebra}\textbf{The making of Fig.~\ref{fig:animal_variations} (bottom).} Exploring possible SR solutions to the LR input, by performing two different editing processes, yielding two different animals.}
\end{figure*}

\begin{figure*}[!t]
\centering
\includegraphics[width=\textwidth]{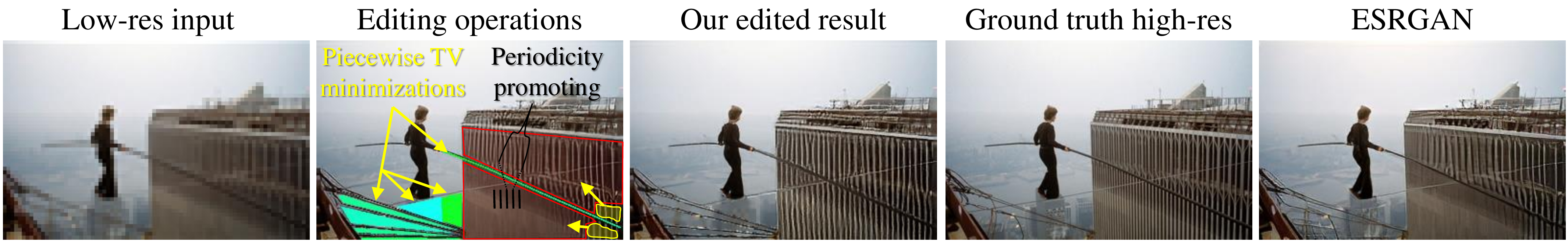}
\caption{\label{fig:editing_rope_woman}\textbf{The making of Fig.~\ref{fig:improved_SR} (top).} Enhancing SR results by utilizing local TV minimization (green-cyan polygons), encouraging periodicity of building patterns and editing distribution of patches - propagating desired patches from source (yellow) to target (red) regions. Corresponding ground truth HR image and result by ESRGAN \cite{wang2018esrgan} are presented for comparison.}
\vspace{-10pt}
\end{figure*}

\begin{figure*}[!t]
\centering
\includegraphics[width=\textwidth]{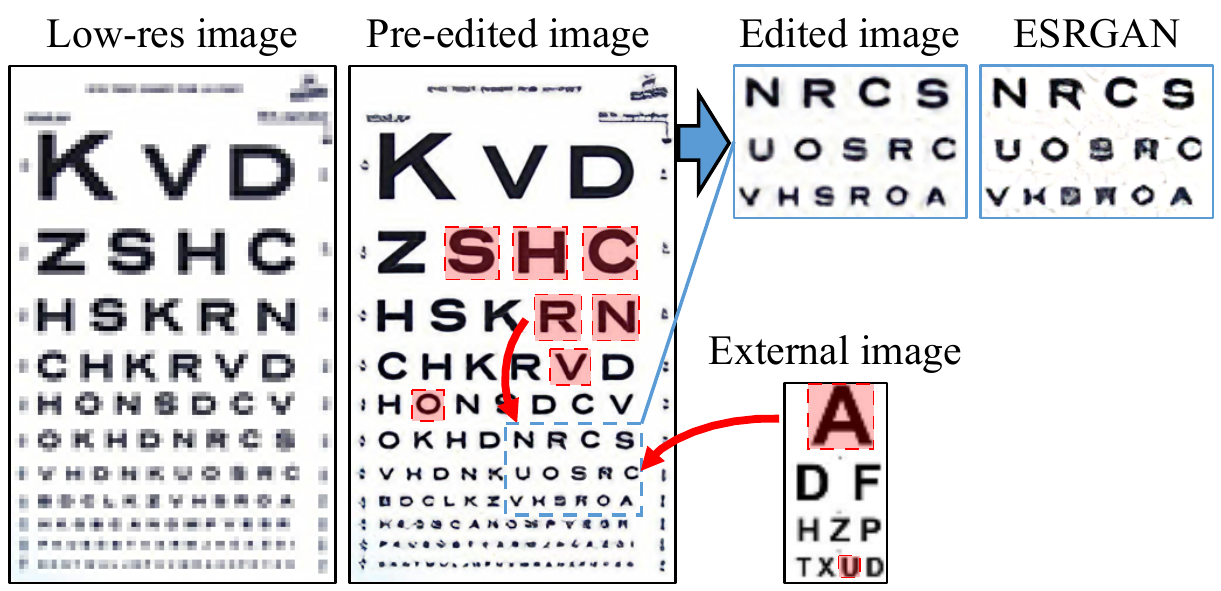}
\caption{\label{fig:across_scales_imprinting}\textbf{Imprinting across scales.} Using our imprinting tool to produce the enhanced SR output presented in Fig.~\ref{fig:improved_SR} (bottom). Most of the imprinted patterns in this example are taken from the same image, exploiting the cross scale recurrence of patterns in natural images, as manifested in this simple case of a Snellen eye-sight chart.}
\vspace{-10pt}
\end{figure*}

We next present editing processes and some additional results obtained using our framework. Figs.~\ref{fig:editing_Bruce} and~\ref{fig:face_editing_Macron_using_Macron} present editing processes and resulting SR outputs for editing face images of Bruce Willis and Emmanuel Macron, respectively. Unlike in the editing process presented in Fig.~\ref{fig:face_editing}, the process in Fig.~\ref{fig:face_editing_Macron_using_Macron} makes use of an external image of Macron himself, yielding enhanced imprinting results. Figs.~\ref{fig:editing_spotted_deer} and~\ref{fig:editing_zebra} present the exploration editing processes yielding the different SR outputs presented in Fig.~\ref{fig:animal_variations}, and Figs.~\ref{fig:editing_rope_woman} and~\ref{fig:across_scales_imprinting} depict the editing processes leading to the enhanced SR images presented in Fig.~\ref{fig:improved_SR}.

{\small
\bibliographystyle{ieee_fullname}
\bibliography{Editable_SR}

\begin{thebibliography}{10}\itemsep=-1pt

\bibitem{agustsson2017div2k}
Eirikur Agustsson and Radu Timofte.
\newblock Ntire 2017 challenge on single image super-resolution: Dataset and
  study.
\newblock In {\em Proceedings of the IEEE Conference on Computer Vision and
  Pattern Recognition Workshops}, pages 126--135, 2017.

\bibitem{bell2019kernel_gan}
Sefi Bell-Kligler, Assaf Shocher, and Michal Irani.
\newblock Blind super-resolution kernel estimation using an internal-gan.
\newblock {\em arXiv preprint arXiv:1909.06581}, 2019.

\bibitem{chen2016infogan}
Xi Chen, Yan Duan, Rein Houthooft, John Schulman, Ilya Sutskever, and Pieter
  Abbeel.
\newblock Infogan: Interpretable representation learning by information
  maximizing generative adversarial nets.
\newblock In {\em Advances in neural information processing systems}, pages
  2172--2180, 2016.

\bibitem{dong2014learning}
Chao Dong, Chen~Change Loy, Kaiming He, and Xiaoou Tang.
\newblock Learning a deep convolutional network for image super-resolution.
\newblock In {\em European conference on computer vision}, pages 184--199.
  Springer, 2014.

\bibitem{goodfellow2014gans}
Ian Goodfellow, Jean Pouget-Abadie, Mehdi Mirza, Bing Xu, David Warde-Farley,
  Sherjil Ozair, Aaron Courville, and Yoshua Bengio.
\newblock Generative adversarial nets.
\newblock In {\em Advances in neural information processing systems}, pages
  2672--2680, 2014.

\bibitem{gulrajani2017wgan_gp}
Ishaan Gulrajani, Faruk Ahmed, Martin Arjovsky, Vincent Dumoulin, and Aaron~C
  Courville.
\newblock Improved training of wasserstein gans.
\newblock In {\em Advances in neural information processing systems}, pages
  5767--5777, 2017.

\bibitem{pix2pix2016}
Phillip Isola, Jun-Yan Zhu, Tinghui Zhou, and Alexei~A Efros.
\newblock Image-to-image translation with conditional adversarial networks.
\newblock {\em arxiv}, 2016.

\bibitem{jolicoeur2018relativistic}
Alexia Jolicoeur-Martineau.
\newblock The relativistic discriminator: a key element missing from standard
  gan.
\newblock {\em arXiv preprint arXiv:1807.00734}, 2018.

\bibitem{karras2019stylegan}
Tero Karras, Samuli Laine, and Timo Aila.
\newblock A style-based generator architecture for generative adversarial
  networks.
\newblock In {\em Proceedings of the IEEE Conference on Computer Vision and
  Pattern Recognition}, pages 4401--4410, 2019.

\bibitem{kim2016accurate}
Jiwon Kim, Jung Kwon~Lee, and Kyoung Mu~Lee.
\newblock Accurate image super-resolution using very deep convolutional
  networks.
\newblock In {\em Proceedings of the IEEE conference on computer vision and
  pattern recognition}, pages 1646--1654, 2016.

\bibitem{kligvasser2018xunit}
Idan Kligvasser, Tamar Rott~Shaham, and Tomer Michaeli.
\newblock xunit: Learning a spatial activation function for efficient image
  restoration.
\newblock In {\em Proceedings of the IEEE Conference on Computer Vision and
  Pattern Recognition}, pages 2433--2442, 2018.

\bibitem{lai2018Lap_srn}
Wei-Sheng Lai, Jia-Bin Huang, Narendra Ahuja, and Ming-Hsuan Yang.
\newblock Fast and accurate image super-resolution with deep laplacian pyramid
  networks.
\newblock {\em IEEE Transactions on Pattern Analysis and Machine Intelligence},
  2018.

\bibitem{ledig2017srgan}
Christian Ledig, Lucas Theis, Ferenc Husz{\'a}r, Jose Caballero, Andrew
  Cunningham, Alejandro Acosta, Andrew Aitken, Alykhan Tejani, Johannes Totz,
  Zehan Wang, et~al.
\newblock Photo-realistic single image super-resolution using a generative
  adversarial network.
\newblock In {\em Proceedings of the IEEE conference on computer vision and
  pattern recognition}, pages 4681--4690, 2017.

\bibitem{lim2017edsr}
Bee Lim, Sanghyun Son, Heewon Kim, Seungjun Nah, and Kyoung Mu~Lee.
\newblock Enhanced deep residual networks for single image super-resolution.
\newblock In {\em Proceedings of the IEEE conference on computer vision and
  pattern recognition workshops}, pages 136--144, 2017.

\bibitem{martin2001bsd100}
David Martin, Charless Fowlkes, Doron Tal, Jitendra Malik, et~al.
\newblock A database of human segmented natural images and its application to
  evaluating segmentation algorithms and measuring ecological statistics.
\newblock In {\em Proceedings of the IEEE International Conference on Computer
  Vision}, 2001.

\bibitem{mathieu2015video_prediction}
Michael Mathieu, Camille Couprie, and Yann LeCun.
\newblock Deep multi-scale video prediction beyond mean square error.
\newblock {\em arXiv preprint arXiv:1511.05440}, 2015.

\bibitem{michaeli2010optimization}
Tomer Michaeli and Yonina~C. Eldar.
\newblock {\em Optimization techniques in modern sampling theory}.
\newblock Cambridge, UK: Cambridge Univ. Press, 2010.

\bibitem{Michaeli2013BlindSR}
Tomer Michaeli and Michal Irani.
\newblock Nonparametric blind super-resolution.
\newblock In {\em Proceedings of the IEEE International Conference on Computer
  Vision}, pages 945--952, 2013.

\bibitem{mittal2012niqe}
Anish Mittal, Rajiv Soundararajan, and Alan~C Bovik.
\newblock Making a “completely blind” image quality analyzer.
\newblock {\em IEEE Signal Processing Letters}, 20(3):209--212, 2012.

\bibitem{navarrete2018multi}
Pablo Navarrete~Michelini, Dan Zhu, and Hanwen Liu.
\newblock Multi--scale recursive and perception--distortion controllable image
  super--resolution.
\newblock In {\em Proceedings of the European Conference on Computer Vision
  Workshops}, 2018.

\bibitem{perarnau2016invertible_editing}
Guim Perarnau, Joost Van De~Weijer, Bogdan Raducanu, and Jose~M {\'A}lvarez.
\newblock Invertible conditional gans for image editing.
\newblock {\em arXiv preprint arXiv:1611.06355}, 2016.

\bibitem{sajjadi2017enhancenet}
Mehdi~SM Sajjadi, Bernhard Scholkopf, and Michael Hirsch.
\newblock Enhancenet: Single image super-resolution through automated texture
  synthesis.
\newblock In {\em Proceedings of the IEEE International Conference on Computer
  Vision}, pages 4491--4500, 2017.

\bibitem{shaham2019singan}
Tamar~Rott Shaham, Tali Dekel, and Tomer Michaeli.
\newblock Singan: Learning a generative model from a single natural image.
\newblock {\em arXiv preprint arXiv:1905.01164}, 2019.

\bibitem{shocher2018zssr}
Assaf Shocher, Nadav Cohen, and Michal Irani.
\newblock “zero-shot” super-resolution using deep internal learning.
\newblock In {\em Proceedings of the IEEE Conference on Computer Vision and
  Pattern Recognition}, pages 3118--3126, 2018.

\bibitem{wang2018realistic_texture}
Xintao Wang, Ke Yu, Chao Dong, and Chen Change~Loy.
\newblock Recovering realistic texture in image super-resolution by deep
  spatial feature transform.
\newblock In {\em Proceedings of the IEEE Conference on Computer Vision and
  Pattern Recognition}, pages 606--615, 2018.

\bibitem{wang2018esrgan}
Xintao Wang, Ke Yu, Shixiang Wu, Jinjin Gu, Yihao Liu, Chao Dong, Yu Qiao, and
  Chen Change~Loy.
\newblock Esrgan: Enhanced super-resolution generative adversarial networks.
\newblock In {\em Proceedings of the European Conference on Computer Vision
  (ECCV)}, pages 0--0, 2018.

\bibitem{wang2018progressive}
Yifan Wang, Federico Perazzi, Brian McWilliams, Alexander Sorkine-Hornung, Olga
  Sorkine-Hornung, and Christopher Schroers.
\newblock A fully progressive approach to single-image super-resolution.
\newblock In {\em Proceedings of the IEEE Conference on Computer Vision and
  Pattern Recognition Workshops}, pages 864--873, 2018.

\bibitem{Xian_2018_texture_editing}
Wenqi Xian, Patsorn Sangkloy, Varun Agrawal, Amit Raj, Jingwan Lu, Chen Fang,
  Fisher Yu, and James Hays.
\newblock Texturegan: Controlling deep image synthesis with texture patches.
\newblock In {\em The IEEE Conference on Computer Vision and Pattern
  Recognition (CVPR)}, June 2018.

\bibitem{Zhang_2018_denseNet_sr}
Yulun Zhang, Yapeng Tian, Yu Kong, Bineng Zhong, and Yun Fu.
\newblock Residual dense network for image super-resolution.
\newblock In {\em The IEEE Conference on Computer Vision and Pattern
  Recognition (CVPR)}, June 2018.

\bibitem{zhu2016gan_editing}
Jun-Yan Zhu, Philipp Kr{\"a}henb{\"u}hl, Eli Shechtman, and Alexei~A Efros.
\newblock Generative visual manipulation on the natural image manifold.
\newblock In {\em European Conference on Computer Vision}, pages 597--613.
  Springer, 2016.

\bibitem{zhu2017toward_multimodal}
Jun-Yan Zhu, Richard Zhang, Deepak Pathak, Trevor Darrell, Alexei~A Efros,
  Oliver Wang, and Eli Shechtman.
\newblock Toward multimodal image-to-image translation.
\newblock In {\em Advances in Neural Information Processing Systems}, pages
  465--476, 2017.

\end{thebibliography}
}

\end{document}